\documentclass{tlp}
\usepackage{amssymb}
\usepackage{tikz}

\usepackage{algpseudocode,algorithm,algorithmicx} 
\algnewcommand\INPUT{\textbf{input: }}
\algnewcommand\OUTPUT{\textbf{output: }}
\algrenewcommand\algorithmicrequire{\textbf{Precondition:}}
\algrenewcommand\algorithmicensure{\textbf{Postcondition:}}
\newcommand*\Let[2]{\State #1 $:=$ #2}
\algdef{SxnE}[sIF]{sIf}{sEndIf}[1]{\algorithmicif\ #1\ \algorithmicthen}

\newtheorem{definition}{Definition} 

\newcommand{\tbeg}{\langle}
\newcommand{\tend}{\rangle}
\newcommand{\entails}{\models}

\newcommand{\lpnot}{\mbox{not}\;\,}

\newcommand{\hif}{\leftarrow}

\newcommand{\aspindent}{\hspace*{.5in}}

\newcommand{\var}{\ensuremath{\mathit{Var}}}
\newcommand{\loc}{\ensuremath{\mathit{Loc}}}
\newcommand{\relc}{\ensuremath{\mathit{Flow}}}
\newcommand{\reld}{\ensuremath{\mathit{Trans}}}
\newcommand{\inv}{\ensuremath{\mathit{I}}}
\newcommand{\Input}{\ensuremath{\mathcal{U}}}
\newcommand{\RealUniverse}{\ensuremath{\mathbb{R}^n}}
\newcommand{\update}{\ensuremath{\xi}}
\newcommand{\guard}{\ensuremath{g}}

\renewcommand{\epsilon}{\ensuremath{\varepsilon}}

\newcommand{\Dom}{\ensuremath{\mathit{Dom}}}
\newcommand{\prob}{\ensuremath{\mathit{Prob}}}
\newcommand{\fs}{\ensuremath{\mathit{Fs}}}
\newcommand{\rs}{\ensuremath{\mathit{Rs}}}
\newcommand{\as}{\ensuremath{\mathit{As}}}
\newcommand{\ps}{\ensuremath{\mathit{Ps}}}
\newcommand{\es}{\ensuremath{\mathit{Es}}}
\newcommand{\arity}{\ensuremath{\mathit{arity}}}
\newcommand{\os}{\ensuremath{\mathit{Os}}}
\newcommand{\init}{\ensuremath{\mathit{Init}}}
\newcommand{\Hinit}{\ensuremath{\mathit{Init}}}

\newcommand{\bi}{\begin{itemize}}
\newcommand{\ei}{\end{itemize}}

\DeclareFontFamily{OT1}{psyr}{}
\DeclareFontShape{OT1}{psyr}{m}{n}{<-> psyr}{}
\def\times{{\fontfamily{psyr}\selectfont\char180}}

\def\rrr#1\\{\par
\medskip\hbox{\vbox{\parindent=2em\hsize=6.12in
\hangindent=4em\hangafter=1#1}}}

\usepackage{color}

\begin{document}

\bibliographystyle{acmtrans}

\long\def\comment#1{}

\title{CASP Solutions for Planning in Hybrid Domains}

\author[M. Balduccini, D. Magazzeni, M. Maratea, E. LeBlanc]
{MARCELLO BALDUCCINI \\ 
Saint Joseph's University, US\\
\email{marcello.balduccini@gmail.com}\\
\and DANIELE MAGAZZENI\\
King's College London, UK\\
\email{daniele.magazzeni@kcl.ac.uk}\\
\and MARCO MARATEA \\
University of Genoa, Italy\\
\email{marco@dibris.unige.it}\\
\and EMILY C. LEBLANC \\
Drexel University, US\\
\email{ecl38@drexel.edu}\\
}

\pagerange{\pageref{firstpage}--\pageref{lastpage}}
\volume{\textbf{10} (3):}
\jdate{March 2002}
\setcounter{page}{1}
\pubyear{2002}

\maketitle

\label{firstpage}

\begin{abstract}
CASP is an extension of ASP that allows for numerical constraints to be added in the rules. PDDL+ is an extension of the PDDL standard language of automated planning for modeling mixed discrete-continuous dynamics. 

In this paper, we present CASP solutions for dealing with PDDL+ problems, i.e., encoding from PDDL+ to CASP, and extensions to the algorithm of the {\sc ezcsp} CASP solver in order to solve CASP programs arising from PDDL+ domains. An experimental analysis, performed on well-known linear and non-linear variants of PDDL+ domains, involving various configurations of the {\sc ezcsp} solver, other CASP solvers, and PDDL+ planners, shows the viability of our solution. 

\noindent
\emph{Under consideration in Theory and Practice of Logic Programming (TPLP).}
\end{abstract}


\section{Introduction}\label{sec:intro}

Constraint Answer Set Programming (CASP)~\cite{bbg05} is an extension of ASP~\cite{gl88,gl91,nie99,bar03,mt99} that makes it possible to add numerical constraints to the rules of ASP programs, thus allowing to represent and reason on infinite state systems. In the automated planning community, the PDDL standard language for domain representation has been extended with constructs for modeling mixed discrete-continuous dynamics, thus with a similar objective of CASP development. The improved language of the
planning community is called PDDL+~\cite{pddl+}. Current CASP algorithms and solvers are either based on an eager approach to CASP solving~\cite{mgz08,os12}, where the numerical constraints are processed within the ASP search, or are based on a lazy approach~\cite{bal09}, where first an ASP solution is found, and then the numerical constraints involved in the ASP solution are checked for consistency by a Constraint Satisfaction Problem (CSP) solver. Independently from the approach used, in order to be able to solve PDDL+ problems with CASP technology, there is the need to both define a suitable encoding from PDDL+ to CASP, and to extend the architecture of current CASP solvers.

In this paper, we present an approach for using CASP solving techniques to find plans for PDDL+ problems. In particular, first we design an encoding from PDDL+ to CASP, where we show how the various constructs and the overall planning problems can be expressed in CASP by building on work on reasoning about action and change, and previous work on ASP encoding of planning problems. Given that there is still no standard language for CASP, we will mainly rely on the language constructs of the {\sc ezcsp} solver. This is not a severe limitation: variants of our encoding for other CASP languages, e.g., {\sc clingon}, can be obtained by syntactic modifications. Second, we present an architecture suitable for finding PDDL+ plans for a PDDL+ problem via its translation to CASP and the use of a CASP solver. 

The proposed encoding and solving techniques follow a discretize-and-validate approach~\cite{upmurphi} in which a discretization is applied during the generation of the CASP encoding. In order to select only those CASP solutions that correspond to valid plans, the basic CASP solving algorithm is extended with a further check that uses the plan validator VAL~\cite{val} to verify the validity of the solution found. If the test fails, another CASP solution is found, until a valid plan has been found. Although the architecture is given for the \textsc{ezcsp} solver, it can be extended to other CASP solvers.
 
The paper contains an extensive experimental analysis on well-known PDDL+ domains.
The analysis includes $(i)$ various configurations of the {\sc ezcsp} solver,  $(ii)$ variants of the proposed encoding, exploiting both domain-specific and domain-independent heuristics, $(iii)$ variants of the encoding for use with the {\sc clingon} CASP solver, studying the ramifications of the added language expressivity of the solver, $(iv)$ state-of-the-art PDDL+ planners dReal and {\sc UPMurphi}, and $(v)$ two planning tasks, i.e., finding a plan at fixed (optimal) time step, and by progressively increasing the maximum time step. The experimental evaluation provides interesting results regarding the performance of CASP-based solutions vs. other PDDL+ planning approaches. It also demonstrates that PDDL+ planning is an excellent source of CASP benchmarks, with results that are sometimes surprising. Overall, the results show that our solution employing {\sc ezcsp} performs well across all domains and reasoning tasks, at the same time being the only solution able to deal with all cases presented. 

The paper is structured as follows. Section~\ref{sec:prel} introduces needed preliminaries about PDDL+, ASP, and CASP. 
%
%
Section~\ref{sec:algo} presents the proposed architecture for planning in hybrid domains and
discusses our encoding from PDDL+ to CASP, which is exemplified in Section~\ref{sec:encoding} by means of a concrete PDDL+ problem. The experimental analysis is provided in Section~\ref{sec:exp}. Section~\ref{sec:extarch} presents a more sophisticated extension of the  solving architecture. The paper ends by discussing related work in Section~\ref{sec:related}, and by drawing conclusions in Section~\ref{sec:concl}.

\section{Preliminaries}\label{sec:prel}


\noindent
In this section, we provide background knowledge on the main topics covered by the paper. We first introduce the domain of hybrid systems, followed by a brief introduction on AI planning and its application to hybrid domains. Finally, we briefly describe ASP, CSP, and CASP. 


Hybrid systems can be described as hybrid automata~\cite{henzinger}, which are finite state automata
extended with continuous variables that evolve over time. 

\begin{definition}[Hybrid Automaton]
\label{def:ha}
  A \emph{hybrid automaton} is a tuple
  $\ensuremath{\mathcal{H}}=(\loc,\var,\Hinit,\relc,\reld,\inv)$, where
  \begin{itemize}
    \item \loc\ is a finite set of locations, $\var=\{x_1, \dots,
    x_n\}$ is a finite set of real-valued variables,
    $\Hinit(\ell)\subseteq\RealUniverse$ is the set of initial values
    for $x_1,\dots,x_n$ for all locations $\ell$.

    \item For each location $\ell$, $\relc(\ell)$ is a relation over
    the variables in \var\ and their derivatives of the form
    \[ \dot{x}(t) = Ax(t)+u(t), u(t) \in \Input, \]

    where $x(t)\in
    \RealUniverse$, $A$ is a real-valued $n\  \mathtt{x}\  n$ matrix, and $\Input
    \subseteq \RealUniverse$ is a closed and bounded convex set.
    
    \item \reld\ is a set of discrete transitions, a discrete
    transition $tr\in\reld$ being a tuple
    $(\ell,\guard,\update,\ell')$ where $\ell$ and $\ell'$ are the
    source and the target locations, respectively, $\guard$ is the
    guard of $tr$ (given as a linear constraint), and $\update$ is the
    update of $tr$ (given by an affine mapping).
    
    \item $\inv(\ell)\subseteq\RealUniverse$ is an invariant for all
    locations $\ell$. 

  \end{itemize}

\end{definition}

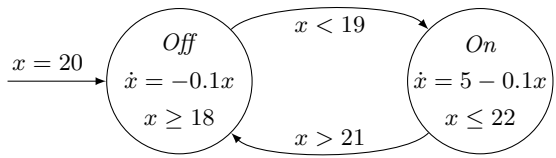
\begin{figure}[h]
  \centering
\begin{tikzpicture}[>=latex]

  \begin{scope}

  
 \node[circle, inner sep=2pt,draw, minimum height=55pt] (p1) at (-4,0) {};
      \node (p0_d) at (-4.0,0.5) {\textit{Off}};
      \node (p0_d) at (-4.0,0) {$\dot{x}=-0.1x$};
      \node (p0_d) at (-4.0,-0.5) {$x \geq 18$};
  
  \node[circle, inner sep=2pt,draw, minimum height=55pt] (p2) at (0,0) {};
      \node (p0_d) at (0.0,0.5) {\textit{On}};
      \node (p0_d) at (0.0,0) {$\dot{x}=5-0.1x$};
      \node (p0_d) at (0.0,-0.5) {$x \leq 22$};
  
     \draw[->] (p1) .. controls +(45:1.5) and +(135:1.5) .. (p2);
     \node (p0_d) at (-2.0,-0.75) {$x > 21$};
     
      \draw[->] (p2) .. controls +(-135:1.5) and +(-45:1.5) .. (p1);
      \node (p0_d) at (-2.0,0.75) {$x < 19$};

      \node (p0t) at (-5.75,0.25) {$x=20$};
      \node (p0) at (-6.40,-0.0) {};
        \draw[->] (p0) edge (p1);  
      
  \end{scope}
\end{tikzpicture}
  \caption{Thermostat hybrid automaton.}
  \label{fig:thermostat}
\end{figure}

An illustrative example is given by the hybrid automaton for a thermostat depicted in Figure~\ref{fig:thermostat}. Here, the temperature is represented by the continuous variable $x$. In the discrete location corresponding to the heater being off, the temperature falls according to the flow condition $\dot{x}=-0.1x$, while when the heater is on, the temperature increases according to the flow condition $\dot{x}=5-0.1x$. The discrete transitions state that the heater \textit{may} be switched on when the temperature falls below 19 degrees, and switched off when the temperature is greater than 21 degrees. Finally, the invariants state that the heater can be on (off) \textit{only} if the temperature is not greater than 22 degrees (not less than 18 degrees).

\subsection{PDDL+ Planning}
Planning is an AI technology that
aims at selecting and organizing activities in order to achieve
specific goals \cite{bookplanning}. A planner uses a domain model, describing
the actions through their pre- and post-conditions,
and an initial state together with a goal condition. It then
searches for a trajectory through the induced state space,
starting at the initial state and ending in a state satisfying the
goal condition. In richer models, such as hybrid systems, the induced state space can
be given a formal semantics as a timed hybrid automaton,
which means that a plan can synchronise activities between
controlled devices and external events. PDDL+ is the planning formalism used to model hybrid system planning domains.\\

\noindent
In the rest of this subsection, we introduce the concepts of planning instance and plan in the context of PDDL+ planning.
 \begin{definition}[Planning Instance]
   A planning instance is a pair $\mathcal{I}=(\Dom,\prob)$, where
   $\Dom=(\fs, \rs, \as,$ $\es, \ps, \arity)$ is a tuple consisting of a
   finite set of \emph{function symbols} \fs, a finite set of
   \emph{relation symbols} \rs, a finite set of (durative)
   \emph{actions} \as, a finite set of \emph{events} \es, a finite set
   of \emph{processes} \ps, and a function \arity\ mapping all symbols
   in $\fs\cup\rs$ to their respective arities.
   The triple $\prob=(\os,\init,G)$ consists of a finite set of
   \emph{domain objects} \os, the \emph{initial} state \init, and the
   \emph{goal} specification $G$.
  
 \end{definition}
Following \cite{bog14}, given a planning instance $\mathcal{I}$, a \textit{state} of $\mathcal{I}$
 consists of a discrete component, described as a set of propositions
 $P$ (the \textit{Boolean fluents}), and a numerical component, described as a set of real
 variables~\textbf{v} (the \textit{numerical fluents}). \emph{Instantaneous} actions are described through
 preconditions (which are conjunctions of propositions in $P$ and/or
 numerical constraints over~\textbf{v}, and define when an action can be
 applied) and effects (which specify how the action modifies the current
 state). The term ``instantaneous'' refers to the fact that instantaneous actions begin and end at the same timepoint. 
Events have preconditions as for
actions, but they are used to model exogenous change in the world;
 therefore they are triggered as soon as the preconditions are true. A
 \emph{process} is responsible for the continuous change of variables, and is
 active as long as its preconditions are true.
 \emph{Durative} actions have three sets of preconditions, representing, respectively,
 the conditions that must hold when an action starts, the invariants that must hold throughout its execution,
 and the conditions that must hold at the end of the action. Similarly, a durative action has three sets of
 effects: effects that are applied when the action starts, effects that are applied when the action ends,
 and a set of continuous numeric effects which are
 applied continuously while the action is executing. 
 
 Formally, the semantics of a PDDL+ planning instance (Definition 2) is given by hybrid automata (Definition 1).
 
The full syntax and semantics of PDDL+ can be found in the seminal paper~\cite{pddl+}. However, in the following we provide a concrete example for the thermostat automaton in Figure \ref{fig:thermostat}, that also shows the main constructs of PDDL+. 

\begin{figure}[h!]
\small
\begin{verbatim}
 (:process off
  :parameters (?t - thermostat ?r - room )
  :condition (and (overall (>= (x ?r) 18)) (isOff ?t))
  :effect
   (and
    (decrease (x ?r) (* #t (* (x ?r) (-0.1)))))
  )
\end{verbatim}
\caption{Structure of a PDDL+ process}
\label{fig:process}
\end{figure}

Figure \ref{fig:process} shows the PDDL+ process representing the continuous change of the temperature when the thermostat is off (that corresponds to the \textit{Off} mode in the hybrid automaton in Figure \ref{fig:thermostat}. Here, the parameters are the thermostat (variable \texttt{?t}) and the room (variable \texttt{?r}). The \texttt{condition} field specifies when the process is active, that is while the thermostat is off and the temperature of the room \texttt{(x ?r)} is greater than or equal to 18. The \texttt{effect} field specifies the continuous change that takes place while the process is active, and actually models the differential equation described in the hybrid automaton of Figure \ref{fig:thermostat}.
A similar process can be defined for when the thermostat is on.
Note that PDDL+ processes follow a \emph{must} semantics, meaning that they are triggered as soon as the preconditions are true, and then become inactive as soon as they become false.

The other key element of a planning domain is the set of actions. Figure \ref{fig:action} shows an example of action for switching the thermostat off.

\begin{figure}[h!]
\small
\begin{verbatim}
 (:action switchOff
  :parameters (?t - thermostat ?r - room )
  :precondition (and (isOn ?t) (> (x ?r) 21)))
  :effect
   (and (isOff ?t) (not (isOn ?t)))
  )
\end{verbatim}
\caption{Structure of a PDDL+ action}
\label{fig:action}
\end{figure}

Here, the preconditions specify that the thermostat can be switched off only if it is currently on and if the temperature of the room is greater than 21. The action has two discrete effects that specify that the thermostat is now off and is no longer on. A similar action can be defined for switching the thermostat on. Note that, as opposed to processes, actions follow a \textit{may} semantics, meaning that the planner may (or may not) decide to apply that action. Clearly, each action can only be applied if the precondition is satisfied.

For completeness, we also provide an example of \textit{durative action} in Figure~\ref{fig:dur-act}, from the mars rover, a typical  domain in the AI planning comunity.

\begin{figure}[h!]
\small
\begin{verbatim}
(:durative-action move
 :parameters (?r - rover ?wp1 ?wp2 - waypoint)
 :duration  (= ?duration  (* 2 (distance ?wp1 ?wp2)) 
 :condition (and (at start (at ?r ?wp1)) 
            (overall (> (batteryLevel ?r) 5)) (overall (visible ?wp1 ?wp2))
 :effect (and (at start (not (at ?r ?wp1)))
              (at end (at ?r ?wp2))      
              (decrease (batteryLevel ?r) (* #t 0.4))
         )
)  
\end{verbatim}
\caption{Structure of a PDDL+ durative action}
\label{fig:dur-act}
\end{figure}

This durative action refers to the rover moving between two waypoints. The \textit{duration} field specifies how long the action takes to complete, and in this case is function of the length of the path to be traversed. The condition field can specify conditions that must hold at start (i.e., before the action can actually start), at end (i.e., when the action terminates) and overall (i.e., throughout the duration of the action). In this example, it is required that at start the rover is in the initial waypoint, and for the whole action the two waypoints are  visible and there is enough battery charge. Similarly, the effects of the action can be applied when the action starts, when it terminates, or they can be continuous while the action is being applied. In this example, when the action starts the rover is no longer at waypoint wp1, it will be at waypoint wp2 when the action terminates, and while the action is being executed the battery level decreases continuously.

\smallskip
\smallskip

Once the planning instance has been defined, a planner can be used to find a \textit{plan}, whose formal definition is given in the following.
 
 \begin{definition}[Plan]
 A \textit{plan} for a planning instance $\mathcal{I}=((\fs, \rs, \as, \es, \ps, \arity),(\os,\init,G))$ is a finite set of triples $(t,a,d)\in\mathbb{R} $\ \times\ $ \as $\ \times\ $\mathbb{R}$, where $t$ is a timepoint, $a$ is an action, and $d$ is the action duration.
 \end{definition}
Note that processes and events do not appear in a plan, as they are not under the direct control of the planner.

\smallskip
\smallskip



\medskip




\smallskip
\smallskip
\subsection{Answer Set Programming} 
Let $\Sigma$ be a signature containing constant, function and
relation symbols. Terms and atoms are formed as in first-order  logic. A literal is
an atom $a$ or its
classical
negation
$\neg a$. 
A \emph{rule} is a statement of the form:
\begin{equation}\label{eq:rule}
h \hif l_1, \ldots, l_m, \lpnot l_{m+1}, \ldots, \lpnot l_n
\end{equation}
where $h$ and $l_i$'s are literals and $\mbox{\emph{not}}$ is the so-called
\emph{default negation}. The intuitive meaning of the rule is that a
reasoner who believes $\{ l_1, \ldots, l_m \}$ and has no reason to believe
$\{l_{m+1}, \ldots, l_n\}$, has to believe $h$.
We call $h$ the \emph{head} of the rule, and
$\{ l_1, \ldots, l_m, \lpnot l_{m+1}, \ldots, \lpnot l_n \}$ the \emph{body}
of the rule. Given a rule $r$, we denote its head and body by $head(r)$ and
$body(r)$, respectively. A rule with an empty body is called a \emph{fact}, and indicates that the head is always true. In that case, the connective $\hif$ is often dropped.

A \emph{program} is a pair $\tbeg \Sigma, \Pi \tend$, where $\Sigma$ is
a signature and $\Pi$ is a set of rules over $\Sigma$. Often we
denote programs by just the second element of the pair, and let the
signature be defined implicitly.

A set $A$ of literals is \emph{consistent} if no two
complementary literals, $a$ and $\neg a$, belong to $A$. 
A literal $l$ is \emph{satisfied}
by a consistent set of literals $A$ (denoted by $A \entails l$) if $l \in A$. 
If $l$ is not satisfied by $A$,
we write $A \not\entails l$. A set $\{ l_1, \ldots, l_k \}$
of literals is satisfied by a set of literals $A$
($A \entails \{ l_1, \ldots, l_k \}$) if each $l_i$ is satisfied by $A$.

Programs not containing default negation are called \emph{definite}.
A consistent set of literals $A$ is \emph{closed} under a definite
program $\Pi$ if, for every rule of the form (\ref{eq:rule})
such that the body of the rule is satisfied by $A$, the head is satisfied by $A$.
This allows us to state the semantics of definite programs.
\begin{definition}
A consistent set of literals $A$ is an \emph{answer set} of definite program
$\Pi$ if $A$ is closed under all the rules of $\Pi$ and $A$ is set-theoretically
minimal among the sets closed under all the rules of $\Pi$.
\end{definition}

To define answer sets of arbitrary programs, we introduce the \emph{reduct} of a program $\Pi$ with respect to a set of
literals $A$, denoted by $\Pi^A$. The reduct is obtained from
$\Pi$ by: {\bf (1)} deleting
every rule $r$ such that $l \in A$ for some expression of
the form $\mathit{not}\ l$ from the body of $r$, and {\bf (2)}
removing
all expressions of the form $\mathit{not}\ l$ from the bodies of the
remaining rules.
The semantics of arbitrary ASP programs can thus be  defined as follows. \begin{definition}
A consistent set of literals $A$ is an \emph{answer set} of program
$\Pi$ if it is an answer set of  $\Pi^A$.
\end{definition}

To simplify the programming task, variables (identifiers with an uppercase initial) are allowed in ASP programs. A rule containing variables (a \emph{non-ground} rule)
is viewed as a shorthand for the set of its \emph{ground instances},
obtained by replacing the variables by all possible ground terms.
Similarly, a non-ground program is viewed as a shorthand for the program
consisting of the ground instances of its rules.

There is also a set of useful constructs, introduced informally in the following.
A rule whose head is empty is called \emph{denial}, and states that its body must not be satisfied. A \emph{choice rule} has a head of the form
\[
c_{l}\  \{ m(\vec{X}) : \Gamma(\vec{X}) \}\  c_u
\]
where $m$ is a relation symbol, $\vec{X}$ is a list of variables, $\Gamma(\vec{X})$ is a set of literals that include variables from $\vec{X}$, and $c_{l}$, $c_{u}$ are non-negative integers. A choice rule intuitively states that, whenever the body is satisfied, the number of literals of the form $m(\vec{X})$ where $\Gamma(\vec{X})$ is satisfied must be between $c_{l}$ and $c_{u}$. If not specified, the values of $c_{l}$ and $c_{u}$ are $0$ and $\infty$, respectively. For example, given a relation $q$ defined by $\{ q(a), q(b) \}$, the rule:
\[
1 \{ p(X) : q(X) \} 2.
\]
intuitively identifies three possible sets of conclusions: $\{ p(a) \}$, $\{ p(b) \}$, and $\{ p(a), p(b) \}$.

\smallskip

\subsection{ Constraint Satisfaction Problems} 
A \emph{Constraint Satisfaction Problem (CSP)} \cite{smi06} is a triple $\tbeg X, D, C \tend$,
where $X=\{x_1, \ldots, x_n \}$ is a set of variables, $D=\{D_1, \ldots, D_n\}$
is a set of domains such that $D_i$ is the domain of variable $x_i$ (i.e.,
the set of possible values that the variable can be assigned), and $C$
is a set of constraints.\footnote{Strictly speaking, the use of
the same index $i$ across sets $X$ and $D$ in the above definition of the set
of domains would require $X$ and $D$ to be ordered. However, as the definition
of CSP is insensitive to the particular ordering chosen,
we follow the approach, common in the literature on constraint satisfaction,
of simply considering $X$ and $D$ as sets and abusing notation slightly in
the definition of CSP.
}
Each constraint $c \in C$ is a pair
$c=\tbeg \sigma, \rho \tend$, where $\sigma$ is a vector of variables and $\rho$
is a subset of the Cartesian product of the domains of such variables.

An \emph{assignment} is a pair $\tbeg x_i, a \tend$, where $a \in D_i$, whose
intuitive meaning is that variable $x_i$ is assigned value $a$. A \emph{compound
assignment} is a set of assignments to distinct variables from $X$. A
\emph{complete assignment} is a compound assignment to all the variables in $X$.
A constraint $\tbeg \sigma, \rho \tend$ specifies the acceptable assignments
for the variables from $\sigma$. We say that such assignments \emph{satisfy}
the constraint. 
\begin{definition}
A \emph{solution} to a CSP $\tbeg X, D, C \tend$ is a complete
assignment satisfying every constraint from $C$.
\end{definition}

Constraints can be represented either \emph{extensionally}, by specifying the pair
$\tbeg \sigma, \rho \tend$, or \emph{intensionally}, by specifying an expression 
involving variables, such as $x < y$. In this paper, we focus on constraints
represented intensionally. A \emph{global constraint} is a constraint that
captures a relation between a non-fixed number of variables \cite{kv06},
such as $sum(x,y,z) < w$ and $all\_dif\!ferent(x_1,\ldots,x_k)$.

One should notice that the mapping of an intensional constraint specification
into a pair $\tbeg \sigma, \rho \tend$ depends on the \emph{constraint domain}.
For example, the expression $1 \leq x < 2$ corresponds to the constraint
$\tbeg \tbeg x \tend, \{ \tbeg 1 \tend \} \tend$ if the finite domain is considered, while
it corresponds to $\tbeg \tbeg x \tend, \{ \tbeg v \tend \,|\, v \in [1,2)  \} \tend$ 
in a continuous domain. For this reason, and in line with
the CLP Schema \cite{jl87,msw06},
in this paper we assume that a CSP includes the specification of the
intended constraint domain.

\smallskip

\subsection{Constraint ASP} 
CASP integrates ASP and CSP in order to deal with continuous dynamics. 
There is currently no standardized definition of CASP, and no stardard language. Multiple definitions have been given in the literature \cite{os12,mgz08,bbg05,bal09}. Although largely overlapping, these definitions and syntax differ from each other in some details.

To ensure generality of our results, in this paper we introduce a simplified definition of CASP, which captures the common traits of the above approaches. The main results will be given using this simplified definition. Later (Section~\ref{sec:encoding}), we will introduce a specific CASP language and use it to describe a practical example. 

\paragraph{Syntax.} In order to accommodate CSP constructs, the language of CASP extends ASP by allowing  \emph{numerical constraints} of the form $x \bowtie y$, where $\bowtie\in\{<,\le,=,\neq, \ge, >\}$, and $x$ and $y$ are \emph{numerical variables}\footnote{Numerical variables are distinct from ASP variables.} or standard arithmetic terms possibly containing numerical variables, numerical constants, and ASP variables. Numerical constraints are only allowed in the head of rules.

\paragraph{Semantics.} Given a numerical constraint $c$, let $\tau(c)$ be a function that maps $c$ to a syntactically legal ASP atom and $\tau^{-1}$ be its inverse. We say that an ASP\ atom $a$ \emph{denotes} a constraint $c$ if $a=\tau(c)$. Function $\tau$ is extended in a natural way to CASP rules and programs. Note that, for every CASP program $\Pi$, $\tau(\Pi)$ is an ASP program. 

Finally, given a set $A$ of ASP literals, let $\gamma(A)$ be the set of ASP\ atoms from $A$ that denote numerical constraints.
%
The semantics of a CASP program can thus be given by defining the notion of CASP solution, as follows.
\begin{definition}\label{def:casp}
A pair $\tbeg A, \alpha \tend$ is a \emph{CASP solution} of 
a CASP program $\Pi$ if-and-only-if $A$ is an answer set of $\tau(\Pi)$ and
$\alpha$ is a solution to $\tau^{-1}(\gamma(A))$.
\end{definition}

\paragraph{{\sc ezcsp} language.} 

In \textsc{ezcsp}, a program is a set of ASP rules written in such a way that their answer sets
encode the desired CSPs. This is accomplished by using three types of special atoms:
\begin{enumerate}
\item
a \emph{constraint domain declaration}, i.e., a statement of the form
$cspdomain(\mathcal{D})$,
where $\mathcal{D}$ is a constraint domain such as fd, q, or r; informally,
the statement says that the CSP is over the specified constraint domain
(finite domains, rational numbers, real numbers),
thereby fixing an interpretation for the intensionally specified constraints of the CSP. A program can contain only one $cspdomain$ declaration;
%
\item
a \emph{constraint variable declaration}, i.e., a statement of the form
$cspvar(x)$,
where $x$ is a ground term denoting a variable of the CSP; 
%
%
\item
a \emph{constraint statement}, i.e. a statement of the form
$required(\gamma)$,
where $\gamma$ is an expression that intensionally represents 
a constraint on (some of) the
variables specified by the $cspvar$ statements;
intuitively, the statement says that the constraint 
represented by $\gamma$ is required to be satisfied
by any solution to the CSP. \\
\end{enumerate}

For example, suppose that we are given integer variables $v(1)$, $v(2)$, $v(3)$ 
and we need to find values for them so that (i) each variable has a distinct value, and (ii) $v(1)+v(2)+v(3) \leq 6$ unless the constraint is blocked (atom $blocked$), in which case the constraint has no effect. A possible \textsc{ezcsp} encoding is: \\
\[
\begin{array}{ll}
d(1).\ \ d(2).\ \ d(3).\ \  \hspace{1cm} cspdomain(\textit{fd}).  \hspace{2cm}
& cspvar(v(X)) \hif d(X).\\
required(v(X) \not= v(Y)) \hif d(X), d(Y), X \not= Y. & required(v(1)+v(2)+v(3) \leq 6) \hif \lpnot blocked.\\
\end{array}
\]
where $X$ and $Y$ are ASP variables used to describe compactly the uniqueness of values of $v(1)$, $v(2)$, $v(3)$.\footnote{For the interested reader, we note that such a constraint can be even more efficiently represented by a global constraint $all\_dif\!ferent$, which is supported by \textsc{ezcsp}.} The grounding of the program is:
\[
\begin{array}{l}
d(1).\ \ d(2).\ \ d(3). \hspace{2.16cm} cspdomain(\textit{fd}).  \\
cspvar(v(1)) \hif d(1).\ \ \ \ \ \ \ \ cspvar(v(2)) \hif d(2).\ \ \ \ \ \ \ \ cspvar(v(3)) \hif d(3).\\
required(v(1) \not= v(2)) \hif d(1), d(2), 1 \not= 2.\ \ \ \ required(v(1) \not= v(3)) \hif d(1), d(3), 1 \not= 3.\ \ \ \ \ldots\\
required(v(1)+v(2)+v(3) \leq 6) \hif \lpnot blocked.\\
\end{array}
\]
%
The answer set of the program is 
$\{ d(1), d(2), cspdomain(fd)$, $cspvar(v(1)),$ $cspvar(v(2))$, $cspvar(v(3))$, $required(v(1) \not= v(2))$, 
$required(v(1) \not= v(3)$, $\ldots$, $required(v(1)+v(2)+v(3) \leq 6) \}$, which intuitively describes the numerical constraints given earlier. Note that constraint $v(1)+v(2)+v(3) \leq 6$ is included in the answer set due to the fact that $blocked$ does not hold.

\section{A CASP-based Solution for Planning in Hydrid Domains}
\label{sec:algo}\label{sec:translation}



%
%
In this section we first present a CASP-based architecture for solving PDDL+ problems. Then, we focus on the CASP encoding of PDDL+ problems.

\subsection{Architecture}

\begin{figure}[htbp]
\begin{center}
\includegraphics[clip=true,trim=0 80 0 0,width=.8\textwidth]{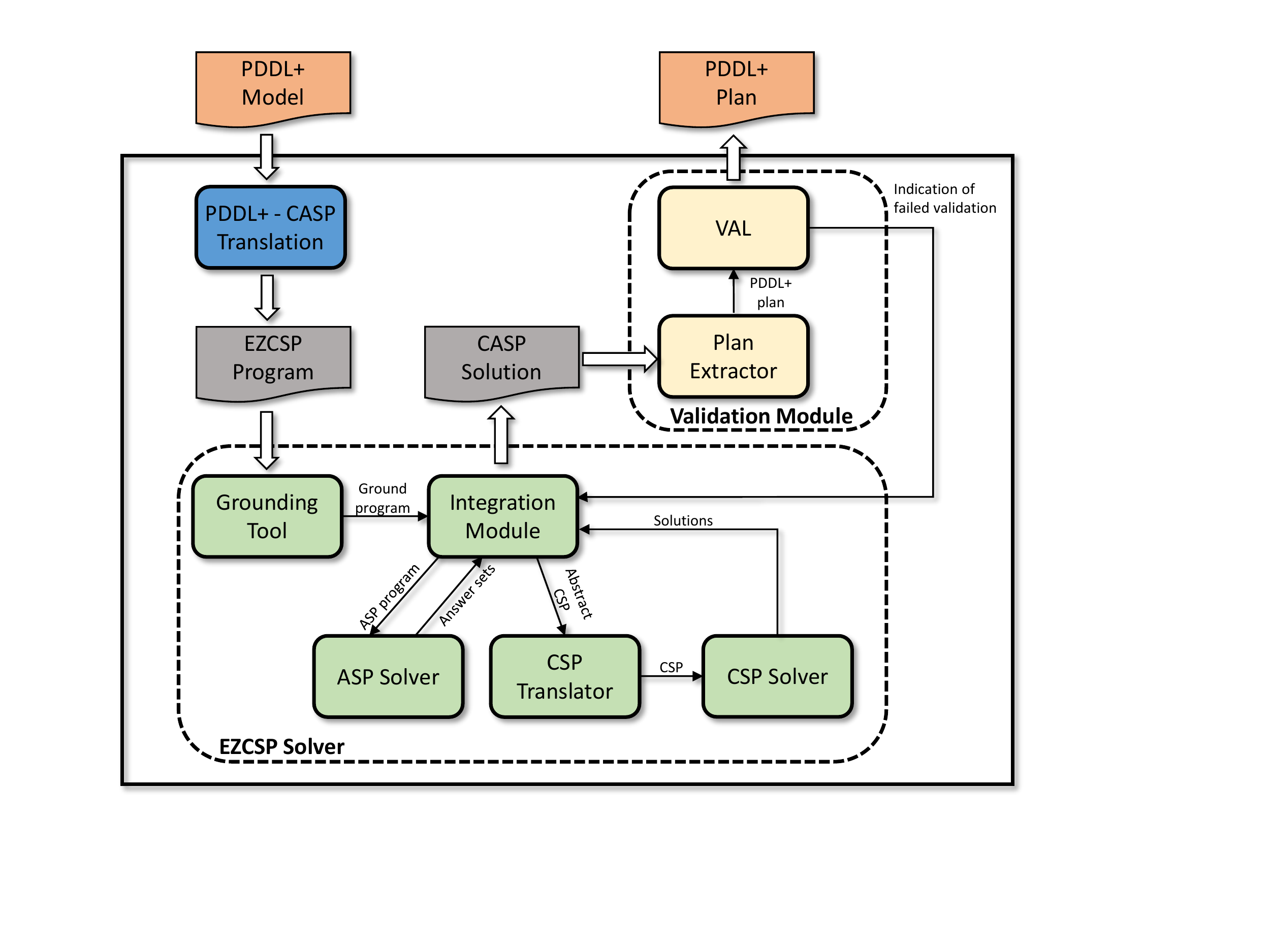}
\end{center}
\caption{CASP-based solution architecture.}\label{fig:ezcsp+val-basic}
\end{figure}

Our approach is centered around the architecture outlined in Figure \ref{fig:ezcsp+val-basic}. 
A PDDL+ model is first translated to a CASP program $\Pi$. Then, the CASP solutions of $\Pi$ are computed by a CASP solver. The architecture is independent of the CASP solver used, but, for presentation purposes, we focus on CASP solver \textsc{ezcsp}. In \textsc{ezcsp}, the Grounding Tool maps $\Pi$ to its syntactically legal ASP counterpart, $\tau(\Pi)$, and grounds it, obtaining program $\Pi_g$. Next, the Integration Module carries out an iterative process in which:
\begin{enumerate}
\item
an ASP solver is used to find an answer set $A$ of $\Pi_g$;
\item
a CSP $C$ is obtained from $A$ by finding all numerical constraints that are denoted by some atom of $A$, i.e.,  $C=\{ \tau^{-1}(a) \,|\, a \in \gamma(A) \}$;
\item
a CSP solver is used to find a solution, $\alpha$, of $C$; 
\item
if $C$ has no 
solution, the ASP solver is used to find a new answer set of $\Pi_g$ and the process is iterated; otherwise, the CASP solution $\tbeg A, \alpha \tend$ is returned. 
\end{enumerate}
More details on the \textsc{ezcsp} algorithm can be found in \cite{bal09}. Details on other CASP solvers can be found in Section~\ref{sec:related}; Section \ref{sec:exp} includes experiments that we conducted using CASP solver \textsc{clingcon}. 

At this point, the Validation Module is executed. The module ensures that the CASP solution found represents a valid PDDL+ plan. As we will see later, in our encoding invariants are enforced at the beginning and at the end of every state in which durative actions and processes are in execution. In this sense, the continuous timeline is discretized. This may cause some plans found by the planner to be invalid: in fact, it is possible for an invariant to be satisfied at the boundaries of each state, and yet be violated within that interval. An example of this behavior is discussed in Section \ref{sec:extarch}. To handle these situations, we follow the discretize-and-validate approach first proposed in \cite{upmurphi}. In that approach, a plan found in the discretized setting is validated using the plan validator VAL~\cite{val}, which is capable of carrying out the calculations needed to check the satisfaction of invariants within state boundaries. The Validation Module consists of two components, the Plan Extractor and VAL. The Plan Extractor component applies syntactic transformations to translate the CASP solution to a PDDL+ plan in the input language of VAL. The plan is then passed to VAL. If VAL finds it to be valid, the plan is returned and the process terminates. Otherwise, \textsc{ezcsp} is invoked again. Suitable constraints are added to the \textsc{ezcsp} program to rule out previously-found CASP solutions and the process is iterated.

\subsection{Encoding}

This section is dedicated to discussing our encoding of PDDL+ problems in CASP.  
Our approach is based on research on reasoning about actions and change, and action languages \cite{gl93,Reiter01,cgw05}. It builds upon the existing SAT-based~\cite{KautzS92} and ASP-based planning approaches and extends them  to hybrid domains.



In reasoning about actions and change, the evolution of a domain over time is often represented by a \emph{transition diagram} (or, \emph{transition system}) that represents states and transitions between states through actions. Traditionally, in transition diagrams, actions are instantaneous, and states have no duration and are described by sets of Boolean fluents. Sequences of states characterizing the evolutions of the domain are represented as a sequence of \emph{discrete time steps}, identified by integer numbers,  so that step 0 corresponds to the initial state in the sequence. We extend this view to hybrid domains according to the following principles:
\begin{itemize}
\item
Similarly to PDDL+, a state is characterized by Boolean fluents and numerical fluents.
\item
The flow of actual time is captured by the notion of \emph{global time} \cite{cgw05}. States have a duration, given by the global time at which a state begins and ends. Intuitively, this conveys the intuition that time flows ``within'' the state.
\item
The truth value of Boolean fluents only changes upon state transitions, that is, it is unaffected by the flow of time ``within'' a state. On the other hand, the value of a numerical fluent may change within a state.
\item The global time at which an action occurs is identified with the end time of the state in which the action occurs. 
\end{itemize}




%
%



Next, we describe the CASP formalization of PDDL+ models. 
We begin by discussing the correspondence between global time and states; then, we present domain, problems and planning task encodings. 

\subsubsection{Time, States, Fluents}
The global time at which the state at step $i$ begins is represented by the numerical variable $tstart(i)$. Similarly, the end time is represented by $tend(i)$. 
%
A fluent literal is a fluent $f$ or its negation $\neg f$. By $\overline{l}$ we denote Boolean complement over fluent literals, i.e., $\overline{f}=\neg f$ and $\overline{\neg f}=f$. If $f$ holds at discrete time step $i$, in the encoding we write $holds(f,i)$; if $\neg f$ holds, i.e., $f$ is false, we write $\neg holds(f,i)$.
We also often use $\chi(l,i)$ as an abbreviation, corresponding to $holds(f,i)$ if $l=f$ and to $\neg holds(f,i)$ if $l=\neg f$. 
%
For every {\sl numerical fluent n}, we introduce two numerical variables representing its value at the beginning and at the end of time step $i$. The variables are $v\_initial(n,i)$ and $v\_final(n,i)$, respectively.
%
The occurrence of an action $a$ at time step $i$ is represented by an atom $occurs(a,i)$.

Additive fluents, whose value is affected by $\mathtt{increase}$ and $\mathtt{decrease}$ statements of PDDL+, are represented by introducing numerical variables of the form
$v(contrib(n,s),i)$, 
where $n$ is a numerical fluent, $s$ is a constant denoting a source (e.g., the action that causes the increase or decrease), and $i$ is a time step. The expression denotes the amount of the contribution to fluent $n$ from source $s$ at step $i$. Intuitively, the value of $n$ at the end of step $i$ (encoded by numerical variable $v\_final(n,i)$) is calculated from the values of the individual contributions.
%
%
%
Next, we discuss the encoding of the domain portion of a PDDL+ problem.

\subsubsection{Domain Encoding}
In the following discussion, ASP variables $I$, $I1$, $I2$ denote time steps.

\medskip
\textbf{Istantaneous Actions.} As we have seen, instantaneous actions are characterized by a set of preconditions and a set of effects. A Boolean precondition is a fluent literal $l$, representing a condition on the truth value of $l$. A numerical precondition is an inequality between mathematical expressions involving combinations of numerical fluents and numerical constants. The encoding of a precondition depends on its type. Thus, every Boolean precondition $l$ of an action $a$ is encoded by a denial: 
\begin{equation}\label{eq:precond-boolean}
\hif \chi(\overline{l},I), occurs(a,I).
\end{equation}
Notice that a separate denial is used for every precondition. 
To illustrate the encoding, consider an example in which an action $refuel\_with(tk1)$, representing the refueling of some machine using refuel tank $tk1$, has a precondition that tank $tk1$ must be available, represented by $avail(tk1)$. The precondition is thus represented by:
\[
\hif \neg holds(avail(tk1),I), occurs(refuel\_with(tk1),I).
\]
A numerical precondition $\gamma$\ is encoded by means of a numerical constraint $\gamma^f$ obtained from $\gamma$ by replacing every occurrence of a numerical fluent $n$ by an expression $v\_final(n,I)$. The latter represents the value of $n$ at the end of time step $I$. Thus, for every precondition $\gamma$ of an action $a$, the encoding includes a rule:
\begin{equation}\label{eq:precond-numerical}
\begin{array}{l}
\gamma^f \hif occurs(a,I).
\end{array}
\end{equation}
Continuing on the previous example, if action $refuel\_with(tk1)$ has a precondition $level(tk1)>0$, intuitively meaning that the tank must not be empty, the precondition is encoded by a rule:
\[
\begin{array}{l}
v\_final(level(tk1),I)>0 \hif occurs(refuel\_with(tk1),I).
\end{array}
\]   
%
Similarly to Boolean preconditions, Boolean effects are represented in PDDL+ by fluent literals. For every Boolean effect $l$ of an action $a$, the encoding includes a rule:
\begin{equation}\label{eq:effect-boolean}
\chi(l,I+1) \hif occurs(a,I).
\end{equation}
The rule states that $l$ is true at the next time step $I+1$ if the action occurs at (the end of) step $I$. For the effects of actions on numerical fluents, we focus our presentation on the assignment of a value to a numerical fluent\footnote{The handling of additive fluents is discussed in the context of durative actions.}, corresponding to PDDL+ expression $\mathtt{(assign\ n\ e)}$ where $n$ is a numerical fluent and $e$ is a mathematical expression possibly including numerical fluents. An effect of this form for an action $a$ is represented by a rule:
\begin{equation}\label{eq:effect-numerical}
\begin{array}{l}
v\_initial(n,I+1)=e' \hif occurs(a,I).
\end{array}
\end{equation}
where $e'$ is obtained from $e$ by replacing every occurrence of a numerical fluent $n'$ by $v\_initial(n',I+1)$.

\medskip
\textbf{Durative actions.} Recall that a durative action is characterized by two sets of conditions: the start conditions, which are analogous to the preconditions of instantaneous actions, and the \emph{invariants}, i.e. conditions that must be true throughout the execution of the action. Syntactically, start conditions and invariants are represented like the preconditions of instantaneous actions, i.e. with fluent literals and numerical inequalities. The effects of a durative action are divided in at-start effects, at-end effects, and continuous effects. The at-start and at-end effects take place at the beginning and at the end of the action and are treated analogously to the effects of instantaneous actions. The continuous effects only apply to numerical fluents and describe a change of value that is a function of the time elapsed since the start of the action. That is, the continuous effects of durative action $d$ can be viewed as a partial function associating to every numerical fluent $n$ affected by $d$ a mathematical expression $e_n(t)$ that is a function of time (and, possibly, of the value of other numerical fluents). This representation makes it possible to capture the behavior of additive fluents. Additionally, the duration of the action is specified by an inequality $\delta \bowtie e$ where $\delta$ represents the duration and $e$ is an mathematical expression possibly including numerical fluents.

In our formalization, a durative action $d$ is encoded by means of two instantaneous actions, $start(d)$ and $end(d)$. The start conditions of $d$ are mapped to preconditions of $start(d)$ and encoded using (\ref{eq:precond-boolean}) and (\ref{eq:precond-numerical}). Note that a durative action typically affects multiple consecutive states. Our encoding introduces a special Boolean fluent $inprogr(d)$ to denote the states during which a durative action $d$ is in progress. How this fluent is made true/false is discussed later. A Boolean invariant, $l$, of a durative action $d$ is encoded by a denial:
\begin{equation}\label{eq:over-all-boolean}
\begin{array}{l}
\hif \chi(\overline{l},I), holds(inprogr(d),I).
\end{array}
\end{equation}
It is worth noticing the similarity with (\ref{eq:precond-boolean}). A numerical invariant $\gamma$ is encoded by means of two rules:
\begin{eqnarray}
\gamma^i \hif holds(inprogr(d),I).\label{eq:over-all-numerical-i} \\
\gamma^f \hif holds(inprogr(d),I).\label{eq:over-all-numerical-f}
\end{eqnarray}
where $\gamma^f$ is defined as for (\ref{eq:precond-numerical}) and $\gamma^i$ is obtained from $\gamma$ by replacing every occurrence of a numerical fluent $n$ by an expression $v\_initial(n,I)$. Note that these rules enforce the invariants only at the beginning and at the end of every state affected by the action. The task of checking whether the conditions hold at other timepoints is discussed later.

The start and end effects of a durative action $d$ can be easily viewed as the effects of the $start(d)$ and $end(d)$ actions and encoded using (\ref{eq:effect-boolean}) and (\ref{eq:effect-numerical}). To account for the special Boolean fluent $inprogr(d)$, in our encoding the start and end effects of $d$ are expanded to include, respectively, $inprogr(d)$ and $\neg inprogr(d)$.

To account for the effects of concurrent actions on additive fluents, our encoding considers separately the contributions from each action by introducing a special numerical variable $v(contrib(n,d),i)$ for every durative action $d$, numerical fluent $n$ affected by $d$, and time step $i$.
Recall that the continuous effect of $d$ on $n$ is described by a mathematical expression $e_n(t)$. This is encoded in CASP by means of a rule:
\begin{equation}\label{eq:contrib}
\begin{array}{l}
v(contrib(n,d),I)=e_n^i \hif holds(inprogr(d),I).
\end{array}
\end{equation}
where $e_n^i$ is obtained from $e_n(t)$ by replacing every occurrence of a numerical fluent $n'$ by $v\_initial(n',I)$ and every occurrence of $t$ by the expression $tend(I)-tstart(I)$.
For example, if a $refuel$ action causes the level of fuel in a tank to increase linearly with time, its effect is encoded by:
\[
\begin{array}{l}
v(contrib(fuel\_level,refuel),I)=1 \cdot(tend(I)-tstart(I)) \hif holds(inprogr(d),I).
\end{array}
\]
The cumulative effects of durative actions are calculated by means of the following rule of the encoding:
\begin{equation}\label{eq:contrib-total}
\begin{array}{l}
v\_final(N,I) = v\_initial(N,I) +\sum_{d\in D} v(contrib(N,d),I). \\
\end{array}
\end{equation}
where $N$ ranges over additive fluents and $D$ over durative actions. Intuitively, for every numerical fluent $n$, the rule accumulates the effects of all durative actions on  it and updates the value of $n$ by the net result of such effects. For instance, in a domain in which $fuel\_level$ is affected, possibly concurrently, by actions $refuel$ and $consume$, their cumulative effects are modeled by:
\[
\begin{array}{l}
v\_final(fuel\_level,I) = v\_initial(fuel\_level,I) +\sum_{d\in \{consume, refuel\}} v(contrib(fuel\_level,d),I). \\
\end{array}
\]
Note that implementing the summation from (\ref{eq:contrib-total}) in an actual CASP language may require the introduction of additional numerical variables and rules because of syntax restrictions of the implemented languages. In the next section, we show how (\ref{eq:contrib-total}) is implemented in the case of \textsc{ezcsp}.

The encoding also includes a rule:
\begin{equation}\label{eq:inter-state-inertia}
\begin{array}{l}
v\_final(N,I) = v\_initial(N,I) \leftarrow \lpnot ab(N,I). \\
\end{array}
\end{equation}
i.e., a default stating that every additive fluent maintains its value unless explicitly changed. Intuitively, atom $ab(n,i)$ states that fluent $n$ is an exception to the default at time step $i$.
Default (\ref{eq:inter-state-inertia}) must be blocked in every state in which actions affect the fluent. This is accomplished by including in the encoding, for every action $d$ affecting a fluent $n$, a rule:
\begin{equation}\label{eq:intra-state-inertia-exception}
ab(n,I) \leftarrow holds(inprogr(d),I).
\end{equation}
Following the semantics of PDDL+, action $end(d)$ is automatically triggered after $start(d)$. For every durative action $d$, the task of finding a time step at which the end action occurs is accomplished by a choice rule:
\begin{equation}\label{eq:end-trigger}
\begin{array}{l}
\hspace*{-.08in}
1\{ occurs(end(d),I2) : I2>I1 \}1 \hif occurs(start(d),I1).
\end{array}
\end{equation}
If time step $i$ is selected for the occurrence of $end(d)$, the timepoint at which the action occurs is given by the value of $tend(i)$. Notice that, unlike the selection of a time step for $end(d)$, there is no need to state explicitly that a value must be selected for $tend(d)$. The semantics of CASP ensures that a value is selected for every numerical variable in the constraint problem.

Finally, the inequality $\delta \bowtie e$ on the duration of a durative action $d$ is encoded by a rule:
\begin{equation}\label{eq:duration}
\begin{array}{l}
(tend(I2)-tend(I1)) \bowtie e^f \hif occurs(end(d),I2), occurs(start(d),I1).
\end{array}
\end{equation}
where $e^f$ is obtained from $e$ by replacing every occurrence of a numerical fluent $n$ by $v\_final(n,I_1)$.

It is worth noting that the encoding extends to supporting multiple occurrences of the same durative action in a natural way. This is accomplished by adding an argument to the name of the action, i.e., instead of writing $d$ we write $d(i)$. The argument represents the time step at which the action starts. For example, (\ref{eq:duration}) becomes:
\[
\begin{array}{l}
(tend(I2)-tend(I1)) \bowtie e^f \hif occurs(end(d(I)),I2), occurs(start(d(I)),I1).
\end{array}
\]
Intuitively, this approach yields multiple, and completely independent, ``copies'' of the durative action, whose effects and termination can be handled accordingly by the encoding presented. For simplicity of presentation, throughout the paper we make the assumption that durative actions occur only once and thus adopt the simpler writing $d$.

\medskip
\textbf{Processes and Events.}
The encoding of processes and events builds upon the approach  outlined above, respectively, for durative and instantaneous actions. That is, an event is encoded using (\ref{eq:precond-boolean}--\ref{eq:effect-numerical}), while a process is encoded using (\ref{eq:over-all-boolean}--\ref{eq:duration}).  However, recall that their triggering is defined by PDDL+'s \textit{must} semantics (see Section \ref{sec:prel}). In our encoding, this is captured by a choice rule combined with numerical constraints. Let $l_1, \ldots, l_k$ be the Boolean preconditions of an event $v$ and $\gamma_1, \ldots, \gamma_m$ be its numerical preconditions. The encoding includes a choice rule\footnote{Due to the syntactic restrictions of CASP solvers, in practice $\gamma_1, \ldots \gamma_m$ are replaced by syntactically correct symbols that are in one-to-one correspondence with them.}:
\begin{equation}\label{eq:must-semantics}
\begin{array}{l}
1\{ occurs(v,I), is\_false(\gamma_1,I), \ldots, is\_false(\gamma_m,I) \}1 \hif \chi(l_1,I), \ldots, \chi(l_k,I).
\end{array}
\end{equation}

Intuitively, when the Boolean conditions of the event are satisfied, the choice rule states that the event will be triggered unless it is inhibited by unsatisfied numerical conditions. Given a numerical condition $\gamma_i$, let $\overline{\gamma_i}$ denote its complement, e.g., $>$ is replaced by $\leq$ and $=$ by $\not=$. This part of the encoding is completed by introducing, for every $\gamma_i$, a rule:
\begin{equation}\label{eq:must-semantics-2}
\begin{array}{l}
\overline{\gamma_i}\ ^f \hif is\_false(\gamma_i,I).
\end{array}
\end{equation}
where operator $^f$ is defined as for (\ref{eq:precond-numerical}). The rule intuitively states that, if a numerical condition is hypothesized to be unsatisfied, then the corresponding numerical fluents must have values that falsify the condition. It is worth observing how the must semantics is obtained: as soon as all numerical conditions are satisfied by the current state $i$, then (\ref{eq:must-semantics-2}) ensures that $is\_false(\gamma_i,i)$ must be false for each of them. If all Boolean conditions are also satisfied, then (\ref{eq:must-semantics}) is forced to make $occurs(v,i)$ true, thus triggering the occurrence of the event. The triggering of a process $p$ is handled in an analogous way, except that $start(p)$ is the event triggered by (\ref{eq:must-semantics}) rather than $v$. Also note that, while one may be tempted to add a rule complementary to (\ref{eq:must-semantics-2}) such as:
\[
\gamma_i^f \hif \lpnot is\_false(\gamma_i,I), occurs(v,I).
\]
this is made unnecessary by the presence of (\ref{eq:precond-numerical}).  To illustrate our approach, consider an example in which a process $generate$ has start preconditions $enabled$ and $fuel\_level>0$. The corresponding encoding is:
\begin{equation}\label{eq:must-semantics-ex}
\begin{array}{l}
1\{ occurs(start(generate),I), is\_false(fuel\_level>0,I) \}1 \hif holds(enabled,I).\\
v\_final(fuel\_level,I) \leq 0 \hif is\_false(fuel\_level>0,I).
\end{array}
\end{equation}

\noindent
The encoding is completed by the domain-independent rules:
%
%
\begin{equation}\label{eq:inertia}
\begin{array}{l}
tstart(I+1) = tend(I).\\
\ \\
v\_initial(N,I+1) = v\_final(N,I).\\
\ \\
holds(F,I+1) \hif
holds(F,I), \lpnot \neg holds(F,I+1).\\
\neg holds(F,I+1) \hif
\neg holds(F,I), \lpnot holds(F,I+1).\\
\end{array}
\end{equation}
The first rule ensures that there are no gaps between the time intervals associated with consecutive states. The other rules handle propagation of fluent values from one state to the next.

\noindent
\subsubsection{Problem Encoding}
The problem portion of the PDDL+ problem is encoded as follows.

\textbf{Initial state.}
The encoding of the initial state consists of a set of rules specifying the values of fluents in $P\cup v$ at step $0$, where $P$ and $v$ are the sets of Boolean and numerical fluents, respectively.

\textbf{Goals.}
The encoding of a goal consists of a set of denials on Boolean fluents and of constraints on numerical fluents, obtained similarly to the encoding of preconditions of actions, discussed earlier.\\

Given a PDDL+ planning instance $\mathcal{I}$, by $\Pi(\mathcal{I})$ we denote the CASP encoding of $\mathcal{I}$. Next, we turn our attention to the planning task.

\noindent
\subsubsection{Planning Task}
Our approach to planning leverages techniques from ASP-based planning \cite{lif02,bgn07}. The planning task is specified by the planning module, $M$, which includes a choice rule of the form:
$$
\lambda\{ occurs(a_1,I), occurs(a_2,I), \ldots, occurs(a_k,I) \} \mu.
$$
where $a_1 \ldots a_k$ are actions as defined by the PDDL+ specification and bounds $\lambda$, $\mu$ allow for control on concurrency of the actions. In practice, planning modules often also include heuristics for the purpose of increasing performance, also in terms of plan quality.

It can be shown that the plans for a given maximum time step for a PDDL+ planning instance $\mathcal{I}$ are in one-to-one correspondence with the CASP solutions of $\Pi(\mathcal{I}) \cup M$ that pass the validation test by VAL. The plan encoded by a CASP solution $A$ can be obtained from the atoms of the form $occurs(a,i)$ and from the value assignments to numerical variables $tstart(i)$ and $tend(i)$.

It is also worth noting the level of modularity of our approach. In particular, it is straightforward to perform other reasoning tasks besides planning (e.g, a hybrid of planning and diagnostics is often useful for applications) by replacing the planning module by a different one, as demonstrated for example in \cite{bg03}. 

%



\section{An Example of our Encoding}\label{sec:encoding}


In this section, we provide an encoding of a PDDL+ instance (a domain and one problem instance that will be used also in the experimental analysis) into a CASP program. 

The specific domain is described in the first subsection, then its PDDL model is given, while the third subsection is devoted to the encoding into CASP.

\subsection{The {\sl generator} domain}

The {\sl generator} domain is well-known across the planning community and has been a testbed for many planners. The problem revolves around refuelling a diesel powered
generator which has to run for a given duration without overflowing or running dry. This problem is interesting because the time at which the refuel action must be applied within the generate action is critical: the fuel level in the tank must not exceed the capacity of the tank so the planner must realize that the refuel action must be delayed at least $X$ time units after the start of the generate action, but must also occur at least $Y$ time units before the end of the generate action to prevent the fuel level falling below zero, where $X$ and $Y$ depend on the instance. 


\subsection{PDDL+ model}
\label{subsec:ex-gen}

The PDDL+ domain of the generator is shown, in slightly simplified form, in Figure \ref{fig:domain}. The \texttt{generate} process models the generator running. Its invariant, specified  by the \texttt{overall} condition, requires the fuel level in the generator to be no less than 0 at all times during the execution of the process. The process has two continuous effects: it decreases the fuel level (the expression \texttt{(* \#t 1)} states that the change is continuous and linear with respect to time) and increases the value of variable \texttt{generator\_time}, which keeps track of how long the generator ran.

\smallskip
Durative action \texttt{refuel} formalizes the use of a tank for refueling the generator. The action has an invariant stating that the fuel level in the generator must not exceed the capacity of the tank, and is required to have a duration of 10 time units.  
The action has the effect of continuously increasing the fuel level of the generator's tank.
Note that the rate at which the generator's tank is refilled is twice as high as the rate at which fuel is consumed. When both actions are being executed, their effects are combined. Given that the duration of the refuel action is fixed, the total contribution of a refuel tank is 20 units of fuel.

\begin{figure}[h!]
\small
\begin{verbatim}
(define (domain linear_generator)
 (:requirements :fluents :durative-actions :duration-inequalities :adl :typing)
 (:types generator)
 (:predicates (generator-ran) (refueling ?g - generator))
 (:functions (fuelLevel ?g - generator) (capacity ?g - generator))

 (:process generate
  :parameters (?g - generator)
  :condition (overall (>= (fuelLevel ?g) 0))
  :effect
   (and
    (decrease (fuelLevel ?g) (* #t 1)) 
    (increase (generator_time ?g) (* #t 1))
    )
  )
 
 (:durative-action refuel
  :parameters (?g - generator ?t - tank)
  :duration  (= ?duration  10) 
  :condition (overall (< (fuelLevel ?g) (capacity ?g)))
  :effect (increase (fuelLevel ?g) (* #t 2))
 )
)  
\end{verbatim}
\caption{PDDL+ domain of the linear generator}
\label{fig:domain}
\end{figure}

\smallskip
The PDDL+ problem models the instance, as shown in Figure \ref{fig:problem}. In this example, we have one tank, the initial fuel level in the generator is 990, and the capacity of the generator is 1000 units. The goal for generator is to run for 1000 time units. 

\begin{figure}[h!]
\small
\begin{verbatim}
(define (problem linear-generator-prob)
        (:domain generator)
        (:objects gen - generator tank1 - tank)
        (:init  
           (= (fuelLevel gen) 990)
           (= (capacity gen) 1000)
           (available tank1))  
        (:goal (= generator_time 1000))
)  
\end{verbatim}
\caption{PDDL+ problem of the linear generator}
\label{fig:problem}
\end{figure}

\subsection{The {\sc ezcsp} encoding}
The encoding assumes the existence of a relation $step(\cdot)$ defining the (finite) range of integers that represent the discrete steps in the evolution of the domain.  Constant $last\_step$ is set to the largest such integer, and determines the maximum number of steps in the plan, provided as an external parameter to the planner.

\paragraph{\bf Domain-independent component.}
We begin the description of the encoding from  its domain-independent portion. Note that \textsc{ezcsp} requires the explicit declaration of all numerical variables. Thus, the first rules we show capture the declarations of the variables that encode the start and end times of states, accompanied by constraints specifying their domain, i.e. they must be non-negative and have values such that the corresponding states have non-negative duration.
\[
\begin{array}{ll}
cspvar(tstart(I)) \hif step(I). \hspace*{.7in}& required(tstart(I) \geq 0) \hif step(I).\\
cspvar(tend(I)) \hif step(I). & required(tend(I) \geq 0) \hif step(I).  \\
\multicolumn{2}{l}{required(tend(I) \geq tstart(I)) \hif step(I).}
\end{array}
\]
The next axioms encode the first and last set of rules from (\ref{eq:inertia}), namely stating that every state ends at the same time in which the next state begins, and that a Boolean fluent maintains its value unless forced to change. For convenience of representation, the remaining rule from (\ref{eq:inertia}) is written in a domain-dependent way, and thus introduced in the next paragraph.
\[
\begin{array}{l}
required(tstart(I2)=tend(I1)) \hif step(I1), step(I2), I2=I1+1.\\
\ \\
holds(F,I2) \hif
        fluent(F),
        step(I1), step(I2), I2=I1+1,
        holds(F,I1),
        not\ \neg holds(F,I2).\\

\neg holds(F,I2) \hif
        fluent(F),
        step(I1), step(I2), I2=I1+1,
        \neg holds(F,I1),
        not\ holds(F,I2).\\
\end{array}
\]

\paragraph{\bf Domain-dependent component.}
At the core of the domain encoding is the formalization of process $generate$, durative action $refuel$, and of how they affect the fuel level in the generator's tank. Let us start with $generate$. 

The generator's fuel level is modeled by additive numerical fluent $fuel\_level$. The next set of rules declares the numerical variables representing the value of the fluent at the beginning and at the end of each state, and sets limits on their values ($tankcap(\cdot)$ is part of CASP\ encoding of the problem instance, and indicates the capacity of the tank):
\[
\begin{array}{l}
cspvar(v\_initial(fuel\_level,I)) \hif step(I). \\
required(v\_initial(fuel\_level,I) \geq 0) \hif step(I). \\
required(v\_initial(fuel\_level,I) \leq TC) \hif step(I), tankcap(TC).\\ \\

cspvar(v\_final(fuel\_level,I)) \hif step(I).\\
required(v\_final(fuel\_level,I) \geq 0) \hif step(I). \\ 
required(v\_final(fuel\_level,I) \leq TC) \hif step(I), tankcap(TC).\\
\end{array}
\]
The next axiom is an instantiation, to numerical fluent $fuel\_level$, of the second rule from (\ref{eq:inertia}) and states that the value of $fuel\_level$ at the end of a state coincides with its value at the beginning of the next.

\[
\begin{array}{l}

required(v\_initial(fuel\_level,I2)=v\_final(fuel\_level,I1)) \hif
        step(I1), step(I2), I2=I1+1.
\end{array}
\]
Next, we turn our attention modeling how the fluent changes over time, beginning with (\ref{eq:contrib-total}). As mentioned earlier, contributions to additive fluents are represented by means of  numerical variables of the form $v(contrib(\cdot),\cdot)$. Because the sum operator from (\ref{eq:contrib-total}) is not supported directly by the syntax of \textsc{ezcsp}, the \textsc{ezcsp} encoding includes auxiliary rules and variables, which  calculate, separately, the total positive and negative contributions from the sum of the relevant contributions. The positive and negative contributions are distinguished by an extra argument ($incr$ or $decr$) in the name of variable $v(contrib(\cdot),\cdot)$.
\[
\begin{array}{l}
cspvar(v(contrib(fuel\_level,decr),I)) \hif step(I).\\
required(v(contrib(fuel\_level,decr),I) \leq 0) \hif step(I).\\
\ \\
decr(I,v(contrib(fuel\_level,decr,D),I)) \hif
        step(I),
        cspvar(v(contrib(fuel\_level,decr,D),I)).\\
required(sum([decr(I)/2],=,v(contrib(fuel\_level,decr),I))) \hif step(I).\\
\ \\
cspvar(v(contrib(fuel\_level,incr),I)) \hif step(I).\\
required(v(contrib(fuel\_level,incr),I) \geq 0) \hif step(I).\\
\ \\
incr(I,v(contrib(fuel\_level,incr,D),I)) \hif
        step(I),
        cspvar(v(contrib(fuel\_level,incr,D),I)).\\
required(sum([incr(I)/2],=,v(contrib(fuel\_level,incr),I))) \hif step(I).
\end{array}
\]
Above, ASP\ variable $D$ ranges over all possible durative actions that may cause a contribution. Numerical variables $v(contrib(fuel\_level,decr),I)$ and $v(contrib(fuel\_level,incr),I)$ capture the negative and positive contributions at a given step. Relations $decr(\cdot,\cdot)$ and $incr(\cdot,\cdot)$ are used to link group the contributions at each time step. For example, a fact $decr(1,v(contrib(fuel\_level,decr,generate),1))$ states that $v(contrib(fuel\_level,decr,generate),1))$ is one of the (negative) contributions to the fuel level at step $1$. The constraint $required(sum([decr(I)/2],=,v(contrib(fuel\_level,decr),I)))$ sums all of the negative contributions for time step $I$ and assigns the total to numerical variable $v(contrib(fuel\_level,decr),I)$. The summation of the positive contributions works in a similar way. The \textsc{ezcsp} counterpart of (\ref{eq:contrib-total}) is a rule stating that the value of $fuel\_level$ is the sum of the total positive and negative contributions:
\[
\begin{array}{l}
required(v\_final(fuel\_level,I) = v\_initial(fuel\_level,I)+v(contrib(fuel\_level,incr),I)\ - \\ \hspace{8.9cm} v(contrib(fuel\_level,decr),I)) \hif
        step(I).
\end{array}
\]
Next, we focus on the $generate$ process. The next rule defines Boolean fluent $inprogr(generate)$, which indicates whether the process is in progress:
\[
\begin{array}{l}
fluent(inprogr(generate)).
\end{array}
\]
Next, we define a numerical variable formalizing $generate$'s contribution to the fuel level of the generator's tank:
\[
\begin{array}{l}
cspvar(v(contrib(fuel\_level,decr,generate),I)) \hif step(I). \\
required(v(contrib(fuel\_level,decr,generate),I)\geq 0) \hif step(I). \\
\end{array}
\]
The contribution provided by $generate$ while the process is in progress is calculated as per (\ref{eq:contrib}):
\[
\begin{array}{l}
required(v(contrib(fuel\_level,decr,generate),I)=1*(tend(I)-tstart(I))) \hif\\
\aspindent step(I), \\
\aspindent holds(inprogr(generate),I).
\end{array}
\]
The next set of rules follows (\ref{eq:inter-state-inertia}) and (\ref{eq:intra-state-inertia-exception}), expressing the fact that, by default, the contribution by $generate$ is $0$. This is elegantly represented by means of default negation:
\[
\begin{array}{l}
ab(contrib(fuel\_level,decr,generate),I) \hif
        step(I),
        holds(inprogr(generate),I).\\
\ \\
required(v(contrib(fuel\_level,decr,generate),I)=0) \hif\\
\aspindent step(I), \\
\aspindent \lpnot ab(contrib(fuel\_level,decr,generate),I).\\
\end{array}
\]
Next, we model the two (instantaneous) actions that trigger the start and the end of the process. The actions are defined by:
\[
\begin{array}{ll}
action(start(generate)). & action(end(generate)).
\end{array}
\]
The following rule is the counterpart of (\ref{eq:end-trigger}) and ensures that action $end(generate)$ is triggered at some time step following $start(generate)$:
\[
\begin{array}{l}
1 \{ occurs(end(generate),I2) : step(I2) : I2 > I1 : I2 < last\_step \}1 \hif
        step(I),
        occurs(start(generate),I1).
\end{array}
\]
%
The next set of rules formalizes the effect of starting and stopping the process. 
\[
\begin{array}{l}
holds(inprogr(generate),I2) \hif
        step(I1), step(I2), I2=I1+1,
        occurs(start(generate),I1).\\
%
\ \\        
\neg holds(inprogr(generate),I2) \hif
    step(I1), step(I2), I2=I1+1,
    occurs(end(generate),I1).\\
\end{array}
\]
The preconditions and the triggering of action $start(generate)$, which follows the \emph{must} semantics, are encoded according to (\ref{eq:precond-numerical}), (\ref{eq:must-semantics}), and (\ref{eq:must-semantics-2}):
\[
\begin{array}{l}
required(v\_final(fuel\_level,I) \geq 0) \hif step(I), occurs(start(generate),I).\\
\ \\
1\{ occurs(start(generate),I), is\_false(fuel\_level \geq 0,I)\ \}1 \hif step(I), I<last\_step, \\
\hspace*{9.85cm} \neg holds(inprogr(generate),I).\\
required(v\_final(fuel\_level,I)<0) \hif step(I), is\_false(fuel\_level \geq 0,I).\\
\end{array}
\]

\noindent
Durative action $refuel$ (instantiated for each refuel tank) is modeled along the same lines. First of all, fluents and numerical variables are defined to represent whether the action is in progress and the action's contribution to the generator's fuel level (variable $TK$ ranges over the refuel tanks from the problem).
\[
\begin{array}{ll}
fluent(inprogr(refuel(TK))) \hif refuel\_tank(TK).\\
\ \\
cspvar(v(contrib(fuel\_level,incr,refuel(TK)),I)) \hif step(I), refuel\_tank(TK). \\
required(v(contrib(fuel\_level,incr,refuel(TK)),I) \geq 0) \hif step(I), refuel\_tank(TK).
\end{array}
\]
The contribution to $fuel\_level$ while $refuel$ is in progress is defined similarly to that of $generate$:
\[
\begin{array}{l}
required(v(contrib(fuel\_level,incr,refuel(TK)),I)=2*(tend(I)-tstart(I))) \hif\\
\hspace*{7cm}step(I), refuel\_tank(TK),\\
\hspace*{7cm}holds(inprogr(refuel(TK)),I).\\
\ \\
ab(contrib(fuel\_level,incr,refuel(TK)),I) \hif
        step(I), refuel\_tank(TK),
        holds(inprogr(refuel(TK)),I).\\
\ \\
required(v(contrib(fuel\_level,incr,refuel(TK)),I)=0) \hif\\
\hspace*{7cm}step(I), refuel\_tank(TK),\\
\hspace{7cm}not\ ab(contrib(fuel\_level,incr,refuel(TK)),I).\\
\end{array}
\]
Next, the instantaneous actions that correspond to the start and to the end of the durative action are introduced, and their effects defined:  
\[
\begin{array}{ll}

action(start(refuel(TK))) \hif refuel\_tank(TK). \\
action(end(refuel(TK))) \hif refuel\_tank(TK). \\
\ \\
%

1\{ occurs(end(refuel(TK)),I2) : step(I2) : I2>I1 : I2 < last\_step \}1 \hif\\
\hspace{9cm}step(I), refuel\_tank(TK),\\
\hspace{9cm}occurs(start(refuel(TK)),I1).\\
\ \\
%


holds(inprogr(refuel(TK)),I2) \hif
        step(I1), step(I2), I2=I1+1, refuel\_tank(TK),\\
        \hspace{5.5cm}occurs(start(refuel(TK)),I1).\\
\ \\
\neg holds(inprogr(refuel(TK)),I2) \hif
        step(I1), step(I2), I2=I1+1, refuel\_tank(TK),\\
        \hspace{5.75cm}occurs(end(refuel(TK)),I1).\\
\end{array}
\]
Finally, the requirement on the duration of $refuel$ from the PDDL+ specification is captured according to (\ref{eq:duration}), i.e. a rule stating that, whenever $refuel$ ends, the time elapsed between its start and its end must equal the duration specified (variable $RT$ represents the duration of the refuel action; for convenience of representation, the duration is parametrized by relation $duration(\cdot,\cdot)$ rather than being expressed directly by a constant):
\[
\begin{array}{l}
required(tend(I2)-tend(I1))=RT) \hif
        step(I1), step(I2), refuel\_tank(TK), \\
        \hspace{6.35cm}duration(refuel(TK),RT),\\
        \hspace{6.35cm}occurs(end(refuel(TK)),I2),occurs(start(refuel(TK)),I1).
%
\end{array}
\]


\textbf{Planning module.} Action selection is achieved by a choice rule stating that refuel actions are allowed to occur at any time: 
\[
0\{ occurs(start(refuel(TK)),I) : step(I) \}1 \hif refuel\_tank(TK).\\
\]
In order to improve performance, the planning module includes a heuristic  saying that at least an action must be executed at every time step:
\[
\begin{array}{l}
some\_action(I) \hif occurs(A,I).\\
\hif step(I), I < last\_step, \lpnot some\_action(I).
\end{array}
\]


\textbf{Problem encoding.}
Relation $step(\cdot)$ is thus defined by:
\[
step(0..last\_step).
\]
The capacity of the generator's tank is given by:
\[
tankcap(1000). 
\]
The set of available fuel tanks and the duration of the $refuel$ action are specified by:
\[
\begin{array}{l}
refuel\_tank(tank1). \\
duration(refuel(TK),10) \hif refuel\_tank(TK). \\
\end{array}
\]
The initial state is specified in the following, with a set of rules that specifies the initial fuel level, states that no process or durative action is initially in progress, and sets the clock to 0:
\[
\begin{array}{l}
required(v\_initial(fuel\_level,0)=990).\\
\neg holds(inprogr(generate),0).\\
\neg holds(inprogr(refuel(TK)),0) \hif refuel\_tank(TK).\\
required(tstart(0)=0).
\end{array}
\]
Finally, the PDDL+ goal $(=\ generator\_time\ 1000)$ is formalized by:
\[
\begin{array}{l}
duration(generate,1000). \\
\ \\
required(tend(I2)-tend(I1)=D) \hif
        step(I1), step(I2), duration(generate,D),\\
\hspace*{2.26in} occurs(end(generate),I2), occurs(start(generate),I1).\\
\end{array}
\]

\textbf{{\sc ezcsp} solution.}
A CASP solution produced by \textsc{ezcsp} for the above encoding of domain and problem is shown next. For clarity, we focus on the restriction of the solution to relation $occurs$ and to numerical variable $tend(\cdot)$:
\[
\begin{array}{l}
occurs(start(generate),0) \\
occurs(start(refuel(tank1)),0) \\
tend(0)=0.000\\
\ \\
occurs(end(refuel(tank1)),1) \\
tend(1)=10.000\\
\ \\
occurs(end(generate),2) \\
tend(2)=1000.000\\
\end{array}
\]
This solution corresponds to the PDDL+ plan in which $generate$ begins immediately (time $0.000$) and lasts for $1000.000$ time units, and refuel begins immediately and lasts for $10.0$ time units:
\[
\begin{array}{lll}
0.000: & generate & [1000.000] \\
0.000: & refuel(tank1) & [10.000] 
\end{array}
\]

\section{Experiments on PDDL+ planning domains}\label{sec:exp}



We performed an empirical evaluation of the run-time performance of our approach. The comparison was with the state-of-the-art PDDL+ planners dReal \cite{bryce} and UPMurphi. 
Although SpaceEx \cite{bog14} is indeed a related approach, it was not included in the comparison because it is focused on proving only plan non-existence. 

\begin{table*}[ht!]
\centering
\begin{tabular}{|l|l|c|c|c|c|c|c|c|c|}\hline
\textbf{Domain} & \textbf{Solver}& \textbf{1} & \textbf{2} & \textbf{3} & \textbf{4} & \textbf{5} & \textbf{6} & \textbf{7} & \textbf{8} \\\hline\hline
Car linear & \textsc{ezcsp} (B-Prolog) & 0.32&0.31&0.32&0.32&0.32&0.30&0.31&0.31\\\hline
           & \textsc{clingcon} & 0.00&0.00&0.00&0.00&0.00&0.00&0.00&0.00\\\hline
           & dReal & 1.11&1.11&1.15&1.14&1.19&1.13&1.14&1.19\\\hline
Car non-linear & \textsc{ezcsp} (GAMS/couenne) &1.41&0.38&0.49&1.1&0.42&0.52&0.43&2.62\\\hline
           & \textsc{ezcsp} (GAMS/knitro) &0.71&0.68&0.29&0.39&0.25&0.25&0.26&0.84\\\hline
           & dReal & 58.21&162.60&-&-&-&-&-&-\\\hline
\end{tabular}
\caption{Car domain with fixed time step. Results in seconds. Problem instances refer to max acceleration.}\label{tab:res-T1-T2}
\end{table*}

\begin{table*}[ht!]
\centering
\begin{tabular}{|l|l|c|c|c|c|c|c|c|c|}\hline
\textbf{Encoding} & \textbf{Solver}& \textbf{1} & \textbf{2} & \textbf{3} & \textbf{4} & \textbf{5} & \textbf{6} & \textbf{7} & \textbf{8} \\\hline\hline
Basic&\textsc{ezcsp} (B-Prolog)&5.82&2.19&41.77&74.51&114.86&424.80&164.95&-\\\hline
&\textsc{clingcon}&16.88&62.62&-&-&-&-&-&52.43\\\hline
&\textsc{clingcon}/Opt&17.15&62.47&51.76&-&-&-&-&-\\\hline
Heuristic&\textsc{ezcsp} (B-Prolog)&0.28&1.03&4.21&7.25&27.08&43.42&54.83&261.89\\\hline
&\textsc{clingcon}&16.89&61.39&49.40&-&-&-&-&-\\\hline
&\textsc{clingcon}/Opt&16.72&61.28&50.91&-&-&-&-&-\\\hline
Estimator&\textsc{ezcsp} (B-Prolog)&0.27&0.73&1.64&25.64&77.38&303.75&-&-\\\hline
&\textsc{clingcon}&25.10&4.87&82.78&-&-&-&-&-\\\hline
&\textsc{clingcon}/Opt&-&-&-&-&-&-&-&-\\\hline
\multicolumn{2}{|c|}{dReal}   &3.73&-&-&-&-&-&-&-\\\hline
\end{tabular}
\caption{Linear variant of generator with fixed time step. Results in seconds. Problem instances refer to number of tanks.}\label{tab:res-T3}
\end{table*}

\begin{table*}[ht!]
\centering
\begin{tabular}{|l|l|c|c|c|c|c|c|c|c|}\hline
\textbf{Encoding} & \textbf{Solver}& \textbf{1} & \textbf{2} & \textbf{3} & \textbf{4} & \textbf{5} & \textbf{6} & \textbf{7} & \textbf{8} \\\hline\hline
Basic&\textsc{ezcsp} (GAMS/conopt)&1.24&3.25&18.49&-&147.94&-&-&*\\\hline
&\textsc{ezcsp} (GAMS/couenne)&0.99&4.4&24.7&-&192.71&-&-&*\\\hline
&\textsc{ezcsp} (GAMS/ipopt)&1.18&6.2&37.54&-&218.8&-&-&*\\\hline
&\textsc{ezcsp} (GAMS/ipopth)&1.1&5.37&30.55&-&-&-&-&*\\\hline
&\textsc{ezcsp} (GAMS/knitro)&0.82&3.62&20.2&-&166.64&-&-&*\\\hline
&\textsc{ezcsp} (GAMS/lindo)&0.78&3.55&19.2&-&150.89&-&-&*\\\hline
&\textsc{ezcsp} (GAMS/lindoglobal)&0.81&3.42&19.14&-&148.08&-&-&*\\\hline
&\textsc{ezcsp} (GAMS/minos)&0.78&3.3&18.27&-&143.19&-&-&*\\\hline
&\textsc{ezcsp} (GAMS/snopt)&0.8&3.41&18.78&-&148.43&-&-&*\\\hline
Heuristic&\textsc{ezcsp} (GAMS/conopt)&0.72&1.62&0.68&1.05&87.95&256.59&238.93&*\\\hline
&\textsc{ezcsp} (GAMS/couenne)&0.32&2.47&1.48&2.05&120.88&329.82&309.68&*\\\hline
&\textsc{ezcsp} (GAMS/ipopt)&0.33&2.88&1.68&1.93&131.95&-&-&*\\\hline
&\textsc{ezcsp} (GAMS/ipopth)&0.33&2.44&1.07&1.59&123.97&-&-&*\\\hline
&\textsc{ezcsp} (GAMS/knitro)&0.31&1.84&0.79&1.13&126.80&449.79&566.50&*\\\hline
&\textsc{ezcsp} (GAMS/lindo)&0.34&1.73&0.81&1.07&127.09&470.66&314.35&*\\\hline
&\textsc{ezcsp} (GAMS/lindoglobal)&0.33&3.67&0.69&1.15&121.79&329.97&280.94&*\\\hline
&\textsc{ezcsp} (GAMS/minos)&0.31&1.71&0.74&1.05&121.27&296.70&294.98&*\\\hline
&\textsc{ezcsp} (GAMS/snopt)&0.29&1.64&0.66&0.98&104.46&-&-&*\\\hline
Estimator&\textsc{EZCSP} (GAMS/conopt)&0.81&1.25&0.49&1.19&93.10&50.50&-&*\\\hline
&\textsc{EZCSP} (GAMS/couenne)&0.33&1.84&0.76&2.65&132.22&73.10&-&*\\\hline
&\textsc{EZCSP} (GAMS/ipopt)&0.38&1.98&0.79&2.20&145.62&-&-&*\\\hline
&\textsc{EZCSP} (GAMS/ipopth)&0.30&1.73&0.65&1.70&134.29&-&-&*\\\hline
&\textsc{EZCSP} (GAMS/knitro)&0.31&1.37&0.55&1.46&112.19&59.41&-&*\\\hline
&\textsc{EZCSP} (GAMS/lindo)&0.30&1.22&0.50&1.12&97.75&53.66&-&*\\\hline
&\textsc{EZCSP} (GAMS/lindoglobal)&0.28&1.14&0.51&1.12&96.20&52.71&-&*\\\hline
&\textsc{EZCSP} (GAMS/minos)&0.34&1.16&0.52&1.16&94.26&51.12&-&*\\\hline
&\textsc{EZCSP} (GAMS/snopt)&0.35&1.21&0.57&1.19&99.59&51.34&-&*\\\hline
\multicolumn{2}{|c|}{dReal}   &8.18&-&-&-&-&-&-&-\\\hline
\end{tabular}
\caption{Non-linear variant of generator with fixed time step. Results in seconds. Problem instances refer to number of tanks.}\label{tab:res-T4}
\end{table*}

\begin{table*}[ht!]
\centering
\begin{tabular}{|l|l|c|c|c|c|c|c|c|c|}\hline
\textbf{Domain} & \textbf{Solver}& \textbf{1} & \textbf{2} & \textbf{3} & \textbf{4} & \textbf{5} & \textbf{6} & \textbf{7} & \textbf{8} \\\hline\hline
Car linear & \textsc{ezcsp} (B-Prolog)&1.01&0.98&1.04&0.99&0.91&0.85&0.88&0.83\\\hline
           & \textsc{clingcon}&0.00&0.00&0.00&0.00&0.00&0.00&0.00&0.00\\\hline
           & UPMurphi&0.40&0.38&0.38&0.38&0.41&0.39&0.40&0.41\\\hline
Car non-linear & \textsc{ezcsp} (GAMS/couenne)&2.59&1.78&1.90&2.50&1.91&2.00&1.84&3.96\\\hline
               & \textsc{ezcsp} (GAMS/knitro)&2.32&1.49&1.14&1.85&1.14&1.18&1.06&2.13\\\hline
               & UPMurphi & 184.88&-&-&-&-&-&-&-\\\hline
\end{tabular}
\caption{Car domain with cumulative times. Results in seconds. Problem instances refer to max acceleration.}\label{tab:res-T5-T6}
\end{table*}

\begin{table*}[ht!]
\centering
\begin{tabular}{|l|l|c|c|c|c|c|c|c|c|}\hline
\textbf{Encoding} & \textbf{Solver}& \textbf{1} & \textbf{2} & \textbf{3} & \textbf{4} & \textbf{5} & \textbf{6} & \textbf{7} & \textbf{8} \\\hline\hline
Basic&\textsc{ezcsp} (B-Prolog)&1.14&2.71&8.56&12.79&25.90&151.94&96.40&279.81\\\hline
&\textsc{clingcon}&17.71&63.75&-&-&-&-&-&117.91\\\hline
&\textsc{clingcon}/Opt&21.41&83.23&-&-&-&-&-&143.95\\\hline
Heuristic&\textsc{ezcsp} (B-Prolog)&0.89&1.92&5.46&9.93&30.79&50.25&67.97&292.22\\\hline
&\textsc{clingcon}&17.30&62.60&50.58&-&-&-&-&-\\\hline
&\textsc{clingcon}/Opt&17.34&62.80&51.43&-&-&-&-&-\\\hline
Estimator&\textsc{ezcsp} (B-Prolog)&0.83&1.55&3.19&26.27&82.32&318.98&-&-\\\hline
&\textsc{clingcon}&31.44&6.51&103.79&-&-&-&-&-\\\hline
&\textsc{clingcon}/Opt&22.83&71.26&-&-&-&-&-&110.70\\\hline
\multicolumn{2}{|c|}{UPMurphi} & 2.02&12.75&91.80&-&-&-&-&-\\\hline
\end{tabular}
\caption{Linear variant of generator with cumulative times. Results in seconds. Problem instances refer to number of refuel tanks.}\label{tab:res-T7}
\end{table*}

\begin{table*}[ht!]
\centering
\begin{tabular}{|l|l|c|c|c|c|c|c|c|c|}\hline
\textbf{Encoding} & \textbf{Solver}& \textbf{1} & \textbf{2} & \textbf{3} & \textbf{4} & \textbf{5} & \textbf{6} & \textbf{7} & \textbf{8} \\\hline\hline
Basic&\textsc{ezcsp} (GAMS/conopt)&2.30&4.36&42.11&-&152.53&-&-&-\\\hline
&\textsc{ezcsp} (GAMS/couenne)&2.79&10.08&81.56&-&387.30&-&-&-\\\hline
&\textsc{ezcsp} (GAMS/ipopt)&2.00&7.58&80.99&-&226.92&-&-&-\\\hline
&\textsc{ezcsp} (GAMS/ipopth)&1.84&6.77&67.04&-&-&-&-&-\\\hline
&\textsc{ezcsp} (GAMS/knitro)&1.52&4.75&46.69&-&175.50&-&-&-\\\hline
&\textsc{ezcsp} (GAMS/lindo)&1.46&4.57&44.07&-&157.69&-&-&-\\\hline
&\textsc{ezcsp} (GAMS/lindoglobal)&1.46&4.50&43.62&-&155.14&-&-&-\\\hline
&\textsc{ezcsp} (GAMS/minos)&1.49&4.50&46.98&-&153.10&-&-&-\\\hline
&\textsc{ezcsp} (GAMS/snopt)&1.49&4.40&41.15&-&156.87&-&-&-\\\hline
Heuristic&\textsc{ezcsp} (GAMS/conopt)&1.44&2.44&13.10&53.70&88.58&267.11&250.03&-\\\hline
&\textsc{ezcsp} (GAMS/couenne)&1.39&5.72&21.07&77.78&244.31&-&-&-\\\hline
&\textsc{ezcsp} (GAMS/ipopt)&0.97&3.77&24.43&78.26&136.53&-&-&-\\\hline
&\textsc{ezcsp} (GAMS/ipopth)&0.93&3.30&20.31&78.17&126.66&-&-&-\\\hline
&\textsc{ezcsp} (GAMS/knitro)&0.89&2.64&15.06&64.25&109.14&318.25&299.07&-\\\hline
&\textsc{ezcsp} (GAMS/lindo)&0.84&2.45&13.20&54.29&89.98&265.14&254.92&-\\\hline
&\textsc{ezcsp} (GAMS/lindoglobal)&0.91&2.74&14.57&58.92&99.72&278.55&266.06&-\\\hline
&\textsc{ezcsp} (GAMS/minos)&0.87&2.36&14.55&56.95&93.78&285.23&259.65&-\\\hline
&\textsc{ezcsp} (GAMS/snopt)&0.92&2.64&14.87&56.22&92.22&271.60&258.83&-\\\hline
Estimator&\textsc{ezcsp} (GAMS/conopt)&0.88&1.89&12.66&54.95&96.47&55.28&-&-\\\hline
&\textsc{ezcsp} (GAMS/couenne)&1.46&5.29&20.99&81.85&261.31&413.99&-&-\\\hline
&\textsc{ezcsp} (GAMS/ipopt)&1.02&2.89&24.96&82.59&153.27&-&-&-\\\hline
&\textsc{ezcsp} (GAMS/ipopth)&1.04&2.68&20.82&82.26&141.18&-&-&-\\\hline
&\textsc{ezcsp} (GAMS/knitro)&0.96&2.17&15.82&72.33&118.26&69.32&-&-\\\hline
&\textsc{ezcsp} (GAMS/lindo)&0.90&2.03&13.94&58.64&101.62&60.48&-&-\\\hline
&\textsc{ezcsp} (GAMS/lindoglobal)&0.89&2.04&13.47&58.65&101.43&59.94&-&-\\\hline
&\textsc{ezcsp} (GAMS/minos)&0.98&2.04&13.54&56.20&97.59&58.03&-&-\\\hline
&\textsc{ezcsp} (GAMS/snopt)&0.85&1.91&13.15&55.46&95.75&57.28&-&-\\\hline
\multicolumn{2}{|c|}{UPMurphi} & -&-&-&-&-&-&-&-\\\hline
\end{tabular}
\caption{Non-linear variant of generator with cumulative times. Results in seconds. Problem instances refer to number of refuel tanks.}\label{tab:res-T8}
\end{table*}
The experimental setup used a virtual machine running in VMWare Workstation 12 on a computer with an i7-4790K CPU at 4.00GHz. The virtual machine was assigned a single core and 4GB RAM. The operating system was Fedora 22 64 bit. The version of \textsc{ezcsp} used was 1.7.4\footnote{http://mbal.tk/ezcsp/}, with gringo 3.0.5\footnote{http://sourceforge.net/projects/potassco/files/gringo/} and clasp 3.1.3\footnote{https://sourceforge.net/projects/potassco/files/clasp/} as grounding tool and ASP solver, and B-Prolog 7.5\footnote{http://www.picat-lang.org/bprolog/} and GAMS 24.5.7\footnote{http://www.gams.com/} as constraint solvers. The former was used for all linear problems and the latter for the non-linear ones. It should be noted that GAMS acts as a front-end to a large set of constraint solvers. The underlying constraint solver to be used by GAMS is selected at run-time. We performed a thorough evaluation of the underlying solvers on the domains they support. 

In an attempt to evaluate the role of the encoding and of the solver in the resulting performance, we also created variants of our encodings, discussed later, suitable for \textsc{clingcon}, and studied their performance. For this part of the experiments, we used \textsc{clingcon} 2.0.3\footnote{https://sourceforge.net/projects/potassco/files/clingcon/}. Note that \textsc{clingcon} supports only integer variables and linear constraints, and was thus applied only to the linear variants of the domains considered. The other systems used were dReal 2.15.11.\footnote{http://dreal.github.io/}, configured as suggested by its authors, and UPMurphi 3.0.2\footnote{https://github.com/gdellapenna/UPMurphi/}.

The experiments were conducted on the linear and non-linear versions of the \textit{generator} and \textit{car} domains. The CASP encodings were created manually as described earlier. Because of the different ways in which dReal and UPMurphi operate, the comparison with dReal was based on finding a single plan with a given maximum time step (called here \emph{fixed-time step} experiments), as discussed in \cite{bryce}. The results are summarized in Tables \ref{tab:res-T1-T2}-\ref{tab:res-T4}. On the other hand, the comparison with UPMurphi was based on the cumulative times for finding a single plan by progressively increasing the maximum time step (referred to as \emph{cumulative-time} experiments). The results are reported in Tables \ref{tab:res-T5-T6}-\ref{tab:res-T8}.

In the tables, entries marked ``-'' indicate a timeout (threshold $600$ sec). Entries marked ``*'' indicate missing entries due to licensing limitations (see below).
Remarkably, we found that, in all instances, VAL determined that the CASP solutions returned represented valid PDDL+ plans. More information on the topic of validation, as well as an example in which validation fails, can be found in Section \ref{sec:extarch}. Next, we discuss the experimental results obtained for each domain.

 

\medskip
\noindent
\textbf{Car.} The version of the car domain we used is the same that was adopted in \cite{bryce}. In this domain, a vehicle needs to travel a certain goal distance from its start position. The vehicle is initially at rest. Two actions allow the vehicle to accelerate and to decelerate. The goal is achieved when the vehicle reaches the desired distance and its speed is $0$. In the linear variant, accelerating increases the velocity by $1$ and decelerating decreases it by $1$. In the non-linear variant, accelerating increases the acceleration by $1$, and similarly for decelerating. The velocity is influenced by the acceleration according to the usual laws of physics. The calculation also takes into account a drag factor equal to $0.1 \cdot v^2$. 
 The instances were obtained by progressively increasing the range of allowed accelerations (velocities in the linear version) from $[-1,1]$ to $[-8,8]$. 
 
 \textsc{ezcsp} was run on the encoding described earlier. The \textsc{clingcon} encoding was obtained by a straightforward syntactic transformation. As illustrated by Tables \ref{tab:res-T1-T2} and \ref{tab:res-T5-T6},  both CASP encodings solved all instances in approximately constant time -- even in the non-linear case --  demonstrating excellent scalability. The experimental evaluation yielded a number of other interesting results, discussed next. We begin from the outcomes of the fixed-time step experiments  (Table \ref{tab:res-T1-T2}), and later address the cumulative-time experiments.\\

\emph{\textsc{ezcsp} vs \textsc{clingcon}.} \textsc{clingcon} outperformed \textsc{ezcsp} in all experiments on the linear variant of \emph{car}, although the small magnitude of both sets of run-times (negligible time for \textsc{clingcon} and about 0.30 sec for \textsc{ezcsp}) prevents general claims. \\

\emph{GAMS solvers.} Given the GAMS limitations in terms of sizes of the formulas that can be analyzed with the free license, only two GAMS solvers can be employed in this domain. Of those, \textsc{knitro}, which implements state-of-the-art interior-point and active-set methods for dealing with non-linear problems, is the best performer.   \\

\emph{\textsc{ezcsp} vs dReal.} \textsc{ezcsp} substantially outperformed dReal: \textsc{ezcsp} solved all non-linear instances in less than one second, while dReal only solved the two smallest instances, with times that peaked at $3$ orders of magnitude larger than the \textsc{ezcsp} times.  \\


\emph{Cumulative times (Table \ref{tab:res-T5-T6}).} As one might expect, given previous results, \textsc{clingcon} outperformed \textsc{ezcsp} on the cumulative-time experiments as well. The absolute times were still   rather small  -- negligible for \textsc{clingcon} and between 0.83 sec and 1.04 sec for \textsc{ezcsp}. Once again, GAMS's \textsc{knitro} solver had the best performance among the GAMS solvers. Finally, the comparison with UPMurphi shows some interesting results. In the linear case, \textsc{ezcsp} is, in fact, about $2$-$3$ times slower than UPMurphi, although the absolute times are small in both cases. On the other hand, \textsc{ezcsp} outperformed UPMurphi by a much larger margin in the non-linear case, with all instances solved in times between 1.78 sec and 3.96 sec, while UPMurphi only solved the first instance, and with a time of 184.88 sec, i.e., nearly $2$ orders of magnitude slower than \textsc{ezcsp}. Reasons for the performance of UPMurphi seem to be $(i)$ the unoptimized implementation of the discretize-and-validate approach, and $(ii)$ the fact that it performs blind search.

\medskip
\noindent
\textbf{Generator.} 
Our encoding models the transfer of liquid according to Torricelli's law:
\[
v=\sqrt{2gh}.
\]
It should be noted that this is different from the approach used in \cite{bryce}, where a simpler,  but less physically accurate model was used. For a fair comparison with \cite{bryce}, the simpler model was used in reproducing the results for dReal. The instances were generated by increasing the number of refuel tanks from 1 to 8. 

The complexity of this domain makes it amenable to studying various optimizations of the CASP encodings, mostly aimed at improving performance of the encoding with respect to the treatment of the must semantics.\footnote{It is not difficult to see, from our presentation of the CASP encoding, that the must semantics may significantly affect performance.} 

In the following, the CASP encoding presented throughout this paper is referred to as ``Basic.''  
The encoding referred to as ``Heuristic'' leverages the observation, based on straightforward considerations on this domain, that the $generate$ process must start at timepoint $0$. Thus, ``Heuristic'' extends ``Basic'' by a single heuristic stating that action $start(generate)$ must occur immediately. The expectation is that ``Heuristic'' will outperform ``Basic'' in most cases. We are also interested in contrasting the effects of this domain-specific, encoding-level heuristic with those of the sophisticated, algorithm-level, and yet domain-independent, heuristics used in dReal. 

The encoding labeled ``Estimator'' takes the observation about the $generate$ process one step further,  replacing the domain-specific heuristic with rules that, in some conditions, can be used to estimate the value of numerical fluents without calling the constraint solver. This is expected to enable earlier pruning of candidate plans directly within the ASP solver, and result in performance potentially comparable with that of ``Heuristic.'' The encoding is based on the following idea. As we have seen, the conditions that trigger a process can be divided in Boolean and numerical. In the CASP encoding, the former are directly listed in the body of choice rule (\ref{eq:must-semantics}), while the latter are captured by numerical constraints, such as (\ref{eq:must-semantics-2}). Because these constraints are checked by the constraint solver, from an algorithmic perspective, a candidate plan that violates (\ref{eq:must-semantics-2}) is detected only after it has been fully computed by the ASP solver and passed to the constraint solver. This may obviously lead to unnecessary computations. A more economical approach seems to be that of checking as many numerical conditions as possible during the ASP computation. To accomplish that, we maintain extra Boolean fluents that capture the value of numerical fluents, i.e., $has\_val(f,v)$ states that numerical fluent $f$ has value $v$. Keeping track of all changes to numerical fluents is obviously unfeasible on the ASP\ side, but it is indeed possible to keep track of direct assignments of values to them. For example, for every rule of the form
\[
v\_initial(f,I)=v \hif \Gamma.
\]    
where $I$ is as above and $v$ is an integer constant\footnote{The approach can be extended to non-integer constants.}, we introduce a rule:
\[
holds(has\_val(f,v),I) \hif \Gamma.
\]    
The value of such fluents is propagated by a rule:
\[
\begin{array}{l}
holds(has\_val(F,V),I+1) \hif holds(has\_val(F,V),I), \lpnot ab(F,I+1).
\end{array}
\]
The last element in the body of the rule enables blocking the propagation when needed. Specifically, for every rule $h(I) \hif \Gamma$ of the original encoding whose head is a numerical constraint over the value of $f$ (at step $I$) that is not in the form of the assignment considered above, we add to the encoding a rule:
\[
ab(f,I) \hif \Gamma.
\]
It is worth noting that, compared to dReal and to ``Heuristic,'' the ``Estimator'' encoding is both encoding-level and domain-independent. Furthermore, while dReal's heuristics are specific to the PDDL+ planning task, this approach is task-independent.

We also studied the performance of \textsc{clingcon}. To do so, we created translations of all of the three encodings, ``Basic,'' ``Heuristic,'' and ``Estimator.'' These translations are labeled ``\textsc{clingcon}'' in the tables. Note that, due to \textsc{clingcon}'s limitations, the ``\textsc{clingcon}/Opt'' translations are applicable only to linear instances.
Another \textsc{clingcon} variant that we considered, labeled ``\textsc{clingcon}/Opt,'' is specifically aimed at leveraging additional features of \textsc{clingcon}. Differently from \textsc{ezcsp}, the language of \textsc{clingcon} allows for numerical constraints in both the head and body of rules. We attempted to take advantage of this and achieve a more compact encoding of the must semantics. We replaced rules (\ref{eq:must-semantics}) and (\ref{eq:must-semantics-2}) by a single rule, in which numerical constraints occur in the body and can directly trigger the process, as long as the Boolean conditions are satisfied. For example, (\ref{eq:must-semantics-ex}) becomes: 
\begin{equation}\label{eq:must-semantics-cl}
\begin{array}{l}
occurs(generate,I) \hif holds(enabled,I), v\_final(fuel\_level,I) > 0.
\end{array}
\end{equation}
In order to understand our interest in ``\textsc{clingcon}/Opt'', note that (\ref{eq:must-semantics-cl}) is a regular rule, rather than a choice rule. With (\ref{eq:must-semantics}) and (\ref{eq:must-semantics-2}), the ASP solver will tend to generate multiple CSPs in an uninformed way, so that each can be checked by the constraint solver. With the adoption of (\ref{eq:must-semantics-cl}), it is conceivable that the triggering of the process may be achieved earlier in the computation and improve performance, thanks to the tighter integration of the ASP and constraint solver in \textsc{clingcon}, whose algorithm is capable of interleaving the processes of solving the qualitative part of the problem and the numerical one.

Next, we discuss the experimental results. As for the \emph{car} domain, we begin by addressing the outcomes of the fixed-time step experiments (Tables \ref{tab:res-T3} and \ref{tab:res-T4}), and later analyze the cumulative-time experiments.\\

\emph{\textsc{ezcsp} vs \textsc{clingcon}.} \textsc{ezcsp} outperformed \textsc{clingcon} in all experiments on the linear variant of \emph{generator}, except for a single outlier (instance 8 of the ``Basic'' encoding). This is a remarkable result, given that in all previous analyses that the authors are aware of, the tighter coupling between ASP solver and constraint solver featured by \textsc{clingcon} led to better performance, see, e.g.,~\cite{BalducciniL13}. The result is surprising in particular because the seemingly strong interdependency between the qualitative and numerical components of \emph{generator} would have been expected to advantage tight coupled solvers. \\

\emph{``\textsc{clingcon}'' vs ``\textsc{clingcon}/Opt.''} Another surprising result comes from the comparison of the ``\textsc{clingcon}'' and ``\textsc{clingcon}/Opt'' translations. Recall that the latter is designed specifically to take advantage of \textsc{clingcon}'s language features and solving algorithm. Yet, ``\textsc{clingcon}/Opt'' did not yield the performance improvements one might have expected. In the case of  ``Heuristic'', ``\textsc{clingcon}/Opt'' and ``\textsc{clingcon}'' had similar performance; for the ``Basic'' encoding, ``\textsc{clingcon}/Opt'' was able to find a  solution for an instance (instance 3) in which ``\textsc{clingcon}'' timed out, but at the same time the former failed to yield a solution for instance 8, which the latter was able to solve; finally, in the case of the ``Estimator'' encoding,  ``\textsc{clingcon}/Opt'' timed out on every instance, yielding substantially worse performance than ``\textsc{clingcon}.''\\

\emph{GAMS solvers.} GAMS's {\sc conopt} solver, which implements a fast method for finding a first feasible solution, and thus can be efficient also on constraint problems without an objective function like ours, exhibited better scalability than all other GAMS solvers.  \\

\emph{Advanced encodings.} Focusing on \textsc{ezcsp}, and on the GAMS {\sc conopt} solver for the non-linear cases, there was no absolute winner among the ``Basic'', ``Heuristic'' and ``Estimator'' encodings. As one might expect, ``Heuristic'' yielded the best performance in many cases. On the other hand, ``Estimator'' was able to tie or beat  ``Heuristic'' in a number of situations -- most notably,  non-linear instance 6. In the linear instances, ``Basic'' had worse performance than ``Estimator'' in most cases,  but scaled better, solving instance 7 while the latter encoding timed out, suggesting a possible negative impact on scalability due to a somewhat larger encoding for ``Estimator''. It is also worth noting that ``Estimator'' surprisingly exhibited worse scalability in the linear case, where 6 instances were solved, than in the non-linear case, where 7 instances were solved, with the $8^{th}$ giving a GAMS's license error rather than a timeout.\\

\emph{\textsc{ezcsp} vs dReal.} Overall, \textsc{ezcsp} substantially outperformed $dReal$. Particularly interesting is the comparison between ``Estimator'' and dReal, since the former uses declarative,  task-independent heuristics, while the latter employs algorithm-level, task-dependent ones. The execution times for \textsc{ezcsp} for this case (Tables \ref{tab:res-T3} and \ref{tab:res-T4}) ranged between 0.27 sec and 303.75 sec for the linear variant, with $6$ instances solved, and between 0.72 sec and 256.59 sec for the non-linear one (with GAMS's {\sc conopt} solver), with $7$ instances solved.  On the other hand, $dReal$ was only able to solve the first instance in both cases (3.73 sec for the linear one and 8.18 sec for the non-linear one).  \\


\emph{Cumulative times (Tables \ref{tab:res-T7} and \ref{tab:res-T8}).} If we contrast the results based on solvers used, \textsc{clingcon} beat \textsc{ezcsp} in  absolute performance on the largest instance (110.70 sec vs 279.81 sec in instance $8$), but lost when overall  scalability is considered: the \textsc{ezcsp} versions of ``Basic'' and ``Estimator'' solve all instances while the \textsc{clingcon} translations solve $3$ instances at best. ``\textsc{clingcon}/Opt'' yielded mixed performance results: while the ``\textsc{clingcon}'' and ``\textsc{clingcon}/Opt'' translations solved the same number of instances overall, the latter often had worse performance -- up to one order of magnitude for instance 2 when using the ``Estimator'' encoding, but typically $20\%$-$30\%$ worse. On the other hand, the ``\textsc{clingcon}/Opt''  translation of the ``Estimator'' encoding was able to solve instance 8. Interestingly, both \textsc{clingcon} translations of the ``Heuristic'' encoding timed out, while the \textsc{clingcon} translations of the ``Basic'' encoding found solutions. This is surprising given that ``Heuristic'', when compared to ``Basic'', introduces heuristic knowledge that simplifies the problem.

Looking at the non-linear variant of the domain, we see that, once again, GAMS's {\sc conopt} solver was the one with best performance among the GAMS solvers. Focusing on \textsc{ezcsp} (and GAMS's {\sc conopt}), ``Basic'' was slightly better than ``Heuristic'' and ``Estimator'' in some cases (e.g., linear instance $8$), but overall lost to both -- especially in the non-linear instances, where ``Basic'' solves $3$ and $2$ fewer instances respectively. Remarkably,  ``Estimator'' featured marginally better performance than ``Heuristic'' on most easy instances (instances $1$-$3$), but did not scale equally well. Finally, UPMurphi had substantially worse performance than \textsc{ezcsp} in all cases: it solved only $3$ linear instances (vs $8$ for \textsc{ezcsp}) and no non-linear instances (vs $7$ for \textsc{ezcsp}\footnote{It should be noted that the experiments on cumulative times do not result in GAMS licensing errors, because the corresponding instances time out before reaching the point where the error would be generated.}). The speedup yielded by \textsc{ezcsp} reached about one order of magnitude before UPMurphi began to time out.




\medskip
\noindent
\textbf{Summary.}
Considering the entire set of experiments, we can draw a number of important conclusions. First of all, the experiments demonstrate that our CASP-based solution substantially outperforms both dReal and UPMurphi, showing better scalability.
\textsc{ezcsp} is also more scalable than \textsc{clingcon} on challenging problems from the {\sl generator} domain. This is a remarkable result, which goes against previous comparisons (see, e.g., \cite{BalducciniL13}) of the performance of tightly coupled hybrid solving algorithms vs their loosely coupled counterparts. Note that such comparisons were conducted on sets of benchmarks different from the ones considered here. This indicates that
 PDDL+ planning problems may be a new and useful class of valuable  benchmarks for the CASP community. 

Looking specifically at encoding variants and the corresponding techniques, the ``Heuristic'' variant of the encoding was overall the best, while "Estimator" worked well in a number of cases, but did not equal its performance. These results corroborate the considerations that led to creation of these variants. Also, ``\textsc{clingcon}/Opt'' failed to yield consistent performance improvements in spite of its leveraging the more advanced features of the \textsc{clingcon} language and solver, which are not available in \textsc{ezcsp}.

Overall, we believe the empirical results demonstrate the promise of our approach. 



\section{Extended Architecture}
\label{sec:extarch}
As we mentioned, validation through VAL always succeeds in the previous experiments. 
In this section, we discuss how the architecture from  Section \ref{sec:algo} can be extended to handle more efficiently cases in which the validation does not succeed. Specifically, we present a variant of the architecture from Figure \ref{fig:ezcsp+val-basic} aimed at making better use of the information provided by VAL in case of a failed validation.

\begin{figure}[htbp]
\begin{center}
\includegraphics[clip=true,trim=0 80 0 0,width=.8\textwidth]{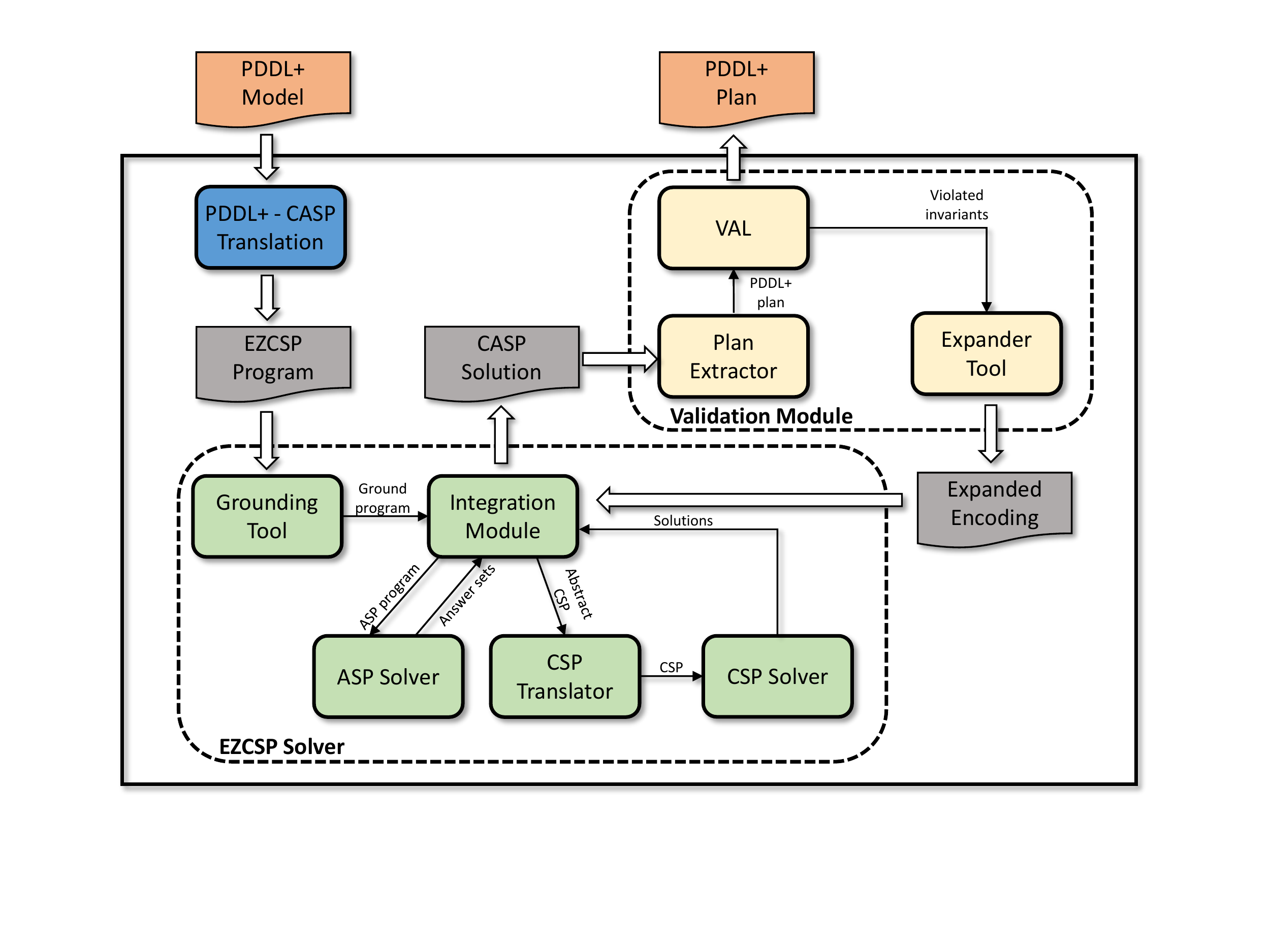}
\end{center}
\caption{Extended {\sc ezcsp} solver architecture.}\label{fig:ezcsp+val}
\end{figure}

The extended architecture is shown in Figure \ref{fig:ezcsp+val}. While the figure refers to \textsc{ezcsp}, the approach is not difficult to extend to \textsc{clingcon} and other CASP solvers.

\begin{figure}[htbp]
\begin{center}
\includegraphics[clip=true,trim=40 250 40 250,width=.8\columnwidth]{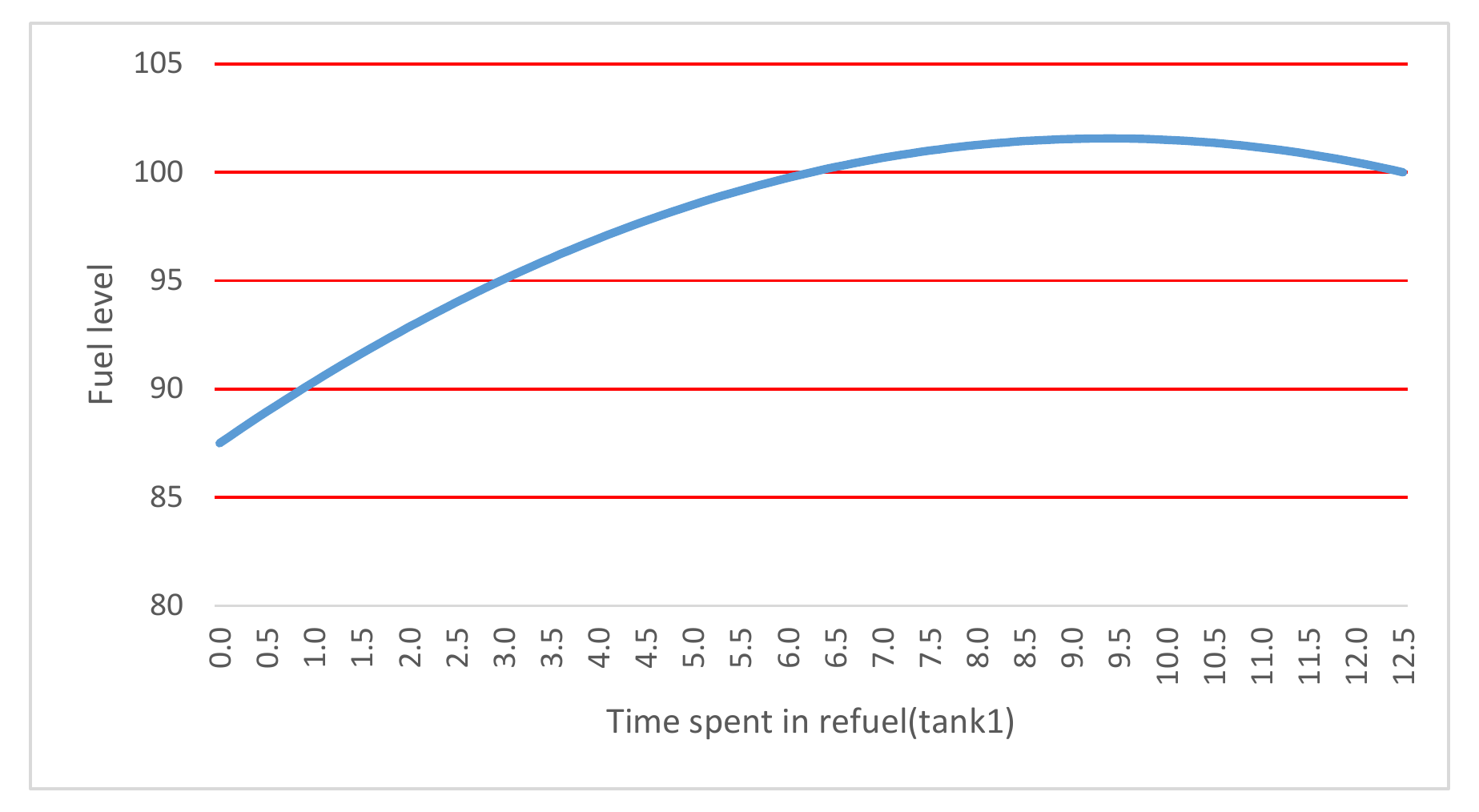}
\end{center}
\caption{Fuel level during refuel(tank1). Capacity 100.00. Spill @ 6.25 sec from the start of the action.}
\label{fig:D-and-V}
\end{figure}

Our approach leverages the fact that, when VAL finds a plan not to be valid, it returns information about which invariants were violated and in which timepoint intervals. Consider a variant of instance 1 of the non-linear generator from Section \ref{sec:exp}, modified so that the capacity and initial level of the generator's tank are $100$ units. A possible plan for this instance is:
\[
\begin{array}{l}
0.000:\ generate\ [100.000]\\
12.500:\ refuel(tank1)\ [12.500]\\
\end{array}
\]
When validating this plan, VAL detects that the invariant
\begin{verbatim}
        (< (fuelLevel ?g) (capacity ?g))
\end{verbatim}
is violated during the execution of $refuel(tank1)$ -- specifically, starting at $6.25$ time units into the execution of the action, corresponding to timepoint range $[18.75,25]$. See Figure \ref{fig:D-and-V} for an illustration.
Such feedback is leveraged by the \emph{Expander Tool} from Figure \ref{fig:ezcsp+val}. The intuition behind the process is to select suitable timepoints in the offending interval, and introduce in the encoding (i) new numerical variables that capture the value of the relevant numerical fluents at those timepoints, and (ii) corresponding numerical constraints that enforce the invariants on those variables. The idea is made precise by Algorithm \ref{alg:expand-main} and Algorithm \ref{alg:expand-aux}. Function \textsc{expand} (Algorithm \ref{alg:expand-main}) takes in input the CASP encoding and a set of pairs $\tbeg inv, range \tend$, where $inv$ is a violated invariant and $range$ is the timepoint range in which it is violated. For each such pair, the algorithm selects a set of timepoints within the range (step \ref{step:expand-main:select}) via user-supplied function \textsc{Select\_Timepoints} and stores them in $\Delta$. These are the timepoints on which the invariant will be enforced by means of additional constraints. In simple cases, \textsc{Select\_Timepoints} may select the timepoints so that they are uniformly distributed within the range, but more complex options are possible. Next (step \ref{step:expand-main:time-step}), the algorithm identifies the time step, $s$, at which the violation of the invariant occurred. For simplicity of presentation, we assume that a single time step corresponds to each range. (If the range extends over multiple time steps, one can always pre-process the output of VAL, splitting each violation into violations whose range corresponds to a single step.) The loop that begins at step \ref{step:expand-main:time-for-1} iterates over every timepoint $\delta$ from $\Delta$ and (step \ref{step:expand-main:inv}) considers each numerical fluent, $n$, that occurs in the invariant. For each of them, $\textsc{Expand\_Fluent}$ (Algorithm \ref{alg:expand-aux}) is called. The function, which will be described in more detail next, expands the encoding with constraints that calculate the value of $n$ at timepoint $\delta$. Next, steps \ref{step:invariant-copy-start}-\ref{step:invariant-copy-end} insert in the encoding a final constraint that instantiates the offending invariant for timepoint $\delta$. At the core of the process is the instantiation of the constraint from the head of the rule of the form (\ref{eq:over-all-numerical-f}) for $inv$, which occurs at step \ref{step:invariant-copy-subst}. In the first part of the process, every occurrence of ASP\ variable $I$ is replaced by the time step, $s$, identified earlier, at which the invariant was violated. Next, for every numerical fluent $n$, the occurrences of $v\_final(n,s)$ are replaced by $v\_final_\delta(n,s)$. This ensures that the constraint considers the values of the fluents at timepoint $\delta$, as calculated by the constraints added by \textsc{Expand\_Fluent}, as opposed to the end of the time step.
\begin{algorithm}
  \begin{algorithmic}[1]
    \Function{Expand}{$\Pi, V$}
    \Statex \INPUT $\Pi$: CASP encoding; $V$: set of pairs $\tbeg inv, range \tend$
    \Let{$\Pi'$}{$\emptyset$}
    \For{$\tbeg inv, range \tend \in V$}
      \Let{$\Delta$}{\textsc{Select\_Timepoints}($range$)}\label{step:expand-main:select} 
      \Let{$s$}{time step at which $inv$ is violated}\label{step:expand-main:time-step}
      \For{$\delta \in \Delta$}\label{step:expand-main:time-for-1}
        \For{$n \in inv$}\label{step:expand-main:inv}
          \Let{$\Pi'$}{$\Pi' \cup \textsc{Expand\_Fluent}(\Pi,n,s,\delta)$}
        \EndFor
        \Let{$d$}{durative action or process for which $inv$ is specified}\label{step:invariant-copy-start}
        \Let{$\gamma^f$}{constraint from the head of the rule of form (\ref{eq:over-all-numerical-f}) for $inv$ }
        \Let{$\gamma^f(s,\delta)$ }{replace in $\gamma^f$:\label{step:invariant-copy-subst}
                  $\left\{\begin{array}{lll}
                    1.\ \ \ I & \mbox{ by } & s \\
                    2.\ \ \ v\_final(n,s) & \mbox{ by } & v\_final_\delta(n,s) \mbox{\ \ \  for every numerical fluent $n$} \\
                   \end{array}\right.$}
        \Let{$\Pi'$}{$\Pi' \cup \{ \gamma^f(s,\delta) \hif holds(inprogr(d),s).\}$}\label{step:invariant-copy-end}
      \EndFor
    \EndFor
    \State \Return{$\Pi'$}
    \EndFunction
  \end{algorithmic}
  \caption{Expansion step}
  \label{alg:expand-main}
\end{algorithm}

Function $\textsc{Expand\_Fluent}$ is given in input the program $\Pi$, a numerical fluent $n$, a timepoint $\delta$ and time step $s$. First of all, the function expands the encoding by creating a rule of the form (\ref{eq:contrib-total}) in which $v\_final(\cdot,\cdot)$ and $v(contrib(\cdot),\cdot)$ are replaced by new variables $v\_final_\delta(\cdot,\cdot)$ and $v_\delta(contrib(\cdot),\cdot).$ These variables capture the value of $n$ at timepoint $\delta$ and the contributions to such value. Next, (steps \ref{step:constraint-copy-start}-\ref{step:constraint-copy-end}) variants of the rules of the form (\ref{eq:contrib}) are introduced. The new rules calculate the contributions to the value of $n$ at timepoint $\delta$ for each durative action and process, $d$. Note that, in rule (\ref{eq:contrib}), such calculation depends on the duration of the action. However, in calculating the contribution of $d$ from its start and up to timepoint $\delta$, the time components of the calculation must be scaled accordingly. This is accomplished by $\textsc{Offset}$ (Algorithm \ref{alg:expand-aux}), which normalizes, w.r.t. $[0,1]$, the position of $\delta$ within the interval of execution of $d$, and by step \ref{step:constraint-scaling} of \textsc{Expand\_Fluent}, which scales the time component accordingly. Recall that $e_n^i$ from (\ref{eq:contrib}) represents the value of the contribution to $n$ from a particular source. Thus, at the core of step \ref{step:constraint-scaling} is the construction of a new expression, $e'$, which represents the contribution up to timepoint $\delta$. Such expression is obtained by instantiating $I$ to the time step at which the invariant was violated and by scaling down, by a factor $\delta_d$, the duration of the interval under consideration.

\begin{algorithm}
  \begin{algorithmic}[1]
    \Function{Expand\_Fluent}{$\Pi, n, s, \delta$}
    \Statex \INPUT $\Pi$: CASP encoding; $n$: numerical fluent; $s$: time step; $\delta$: timepoint
    \Let{$\Pi'$}{$\{ v\_final_\delta(n,s)=v\_initial(n,s)+\sum_{d\in D} v_\delta(contrib(n,d),s). \}$}
    \For{durative action or process $d$}\label{step:constraint-copy-start}
      \For{every rule of $\Pi$ of the form 
      $
      v(contrib(n,d),I)=e_n^i \hif holds(inprogr(d),I)
      $
      }
        \Let{$\delta_d$}{$\textsc{Offset}(\delta,d)$}
        \Let{$e'$}{replace in $e^i_n$:\label{step:constraint-scaling}
                $\left\{\begin{array}{lll}
                  1.\ \ \ I & \mbox{ by } & s \\
                  2.\ \ \ tend(s)-tstart(s) & \mbox{ by } & \delta_d \cdot (tend(s)-tstart(s)) \\
                 \end{array}\right.$}
        \Let{$\Pi'$}{$\Pi' \cup \{ v_\delta(contrib(n,d),I)=e' \hif holds(inprogr(d),s). \}$}\label{step:constraint-copy-end}
      \EndFor
    \EndFor
    \State \Return{$\Pi'$}
    \EndFunction

\algdef{SxnE}[sIF]{sIf}{sEndIf}[1]{\algorithmicif\ #1\ \algorithmicthen}

    \Statex
    \Function{Offset}{$\delta,d$}
    \Statex \INPUT $\delta$: timepoint; $d$: durative action or process
    \Let{$\tbeg s, e \tend$}{start, end timepoints of $d$ in the plan}
    \sIf{$\delta > e$}
      \Return{$1$}
    \sEndIf
    \sIf{$\delta < s$}
      \Return{$0$}
    \sEndIf
    \State \Return $\frac{\delta-s}{e-s}$
    \EndFunction
  \end{algorithmic}
  \caption{Expansion of a numerical fluent}
  \label{alg:expand-aux}
\end{algorithm}

Next, we illustrate the algorithms with an example, which we carry out using the language of \textsc{ezcsp}. Continuing with the output of VAL described earlier, \textsc{Expand} begins by selecting timepoints at which the invariant should be further checked. For sake of illustration, let us select $3$ uniformly distributed timepoints, $18.75$, $21.875$, and $25$. The expansion of the encoding for the negative contribution and timepoint $18.75$ is (below, $\iota(t)$ is the formula that calculates fuel transferred for a duration of time $t$; the actual calculation is omitted for space considerations):
%
\[
\begin{array}{l}
required(
v_{18.75}(contrib(fuel\_level,incr,refuel(TK)),2) = \iota(\frac{18.75-12.5}{25-12.5}*(tend(2)-tstart(2)))) \hif \\
\hspace*{10cm}holds(inprog(refuel(TK)),2). \\
\ \\
required(v\_final_{18.75}(fuel\_level,2) = v\_initial(fuel\_level,2)\\
\hspace{5.6cm} +v_{18.75}(contrib(fuel\_level,incr),2)\\
\hspace{5.6cm} -v_{18.75}(contrib(fuel\_level,decr),2)) \hif
        step(I).\\
\ \\
required(v\_final_{18.75}(fuel\_level,2) \leq TC) \hif tankcap(TC), holds(inprog(refuel(TK)),2).\\
\end{array}
\]
%
The expansion for the positive contributions to the fuel level and for timepoints $21.875$ and $25$ is similar. Once the encoding has been expanded, \textsc{ezcsp} is executed again. The added constraints ensure that any solution that is returned does not violate the  invariants at those timepoints. The process is iterated as needed until the validation succeeds. In our example, for the expanded encoding described above, the planner returns the plan:
\[
\begin{array}{l}
0.000:\ generate\ [100.000] \\
14.063:\ refuel(tank1)\ [12.500]
\end{array}
\]
This plan is successfully validated by VAL, and the search terminates.

The approach has been implemented in prototypical form. When applied to the problem above, the prototype correctly detects that the original plan is
not validated by VAL. Then, using VAL's output, the expanded encoding is generated as above,
and \textsc{ezcsp} is used to find a new solution. Using the ``Heuristic'' encoding, \textsc{ezcsp}, and GAMS's \textsc{conopt} solver, the computation of the first plan takes $0.37$ sec and that of the second plan takes $0.40$ sec, for a total of $0.77$ sec. The time taken for the validation by VAL is negligible. For ``Basic'' and ``Estimator,'' the total times are $1.08$ sec and $0.84$ sec, respectively.


\section{Related Work}\label{sec:related}


In this section, we analyze related approaches and solvers. We start with PDDL+, then we move to CASP. 
Last, we relate our approach to research on planning in hybrid domains based on action languages.  

\paragraph{\bf PDDL+ algorithms and solvers. }
Various techniques and tools have been proposed to deal
 with hybrid domains \cite{zeno,optop,kongming,coles-et-al2012,TM-LPSAT}.
 Nevertheless, none of these approaches are able to handle the full set of PDDL+ features, namely, non-linear domains with processes and events. 
 
More recent works include \cite{bryce}, which presents an approach based on SMT for handling hybrid domains. However, dReach does not use PDDL+, and cannot handle exogenous events.

From the model checking and control communities, a number of works based on timed and hybrid automata have been proposed to handle hybrid systems. Some examples include \cite{hycomp,nuxmv,pappas,maly,hscc1,hscc2,hscc3}, sampling-based planners \cite{rrt,sampl}. Similarly,  \textit{falsification} of hybrid systems tries to guide the search towards the error states, that can be easily cast as a planning problem, \cite{falsif,cimatti1997planning}.
However, while all these works aim to address hybrid systems, they cannot be used to handle PDDL+ models. Some recent works \cite{bog14,bog15} are trying to define a formal translation between PDDL+ and standard hybrid automata, but so far only an over-approximation has been defined, that allows the use of those tools only for proving plan non-existence.

UPMurphi~\cite{upmurphi,upmurphijournal} is the only tool able to handle the full set of PDDL+ features, although it is very limited in scalability as it performs blind search.

PDDL+ has been used to model a number of planning applications \cite{traffic,chemical,batteryICAPS} but then specific domain dependent heuristics or tools have been used to find plans \cite{ucp,batteryJAIR}.

The approach proposed in this paper is similar to the encoding used in TM-LPSAT~\cite{TM-LPSAT}. However, TM-LPSAT assumes linear continuous change, given it uses a linear solver to manage the continuous constraints, while we are solving problems with non-linear dynamics. TM-LPSAT exploits the linearity by checking conditions only at the end-points of intervals of continuous change. Moreover, as reported by Shin and Davies, the experiments showed that TM-LPSAT was not performant and finally the code is not available, hence TM-LPSAT remains a largely theoretical (though valuable) contribution. 

Finally, some very recent works on PDDL+ planning have been proposed ~\cite{smtplan+,dino}, but a proper comparison with them will be addressed in future work.\footnote{At the time of submission, these papers were not published yet, and the corresponding planners not released.}

\paragraph{\bf CASP algorithms and solvers. }
For what concerns CASP solvers, although other CASP solvers exist, \textsc{ezcsp} is, to the best of our knowledge, the only one supporting both real numbers and non-linear constraints, required for modeling non-linear continuous change.

{\sc ACsolver} \cite{mgz08} implements an eager approach to CASP solving, where (in contrast to the lazy approach of {\sc ezcsp}) ASP and CSP solving are tightly coupled and interleaved. It does not support non-linear or global constraints, but allows for real numbers.

{\sc clingon} \cite{os12} is another tightly coupled CASP solver.  The available implementation, however, is not broadly applicable to the kinds of problems considered in this paper. In fact, {\sc clingon} does not support non-linear constraints and real numbers. On the other hand, unlike {\sc ezcsp}, it allows for numerical constraints both in the head of rules and in their bodies, and is characterized by a tighter coupling of ASP and CSP solvers.

A high level view  of the languages and solving techniques employed by these solvers can be found in \cite{Lierler14}. There, by relying on the framework of {\sl abstract solvers}, i.e., a graph-based representation of solving algorithms, similarities and differences among these solvers are formally stated by means of comparison of the related graphs.


\paragraph{\bf Hybrid domains and action languages. }
Action language {\sl H} was introduced in~\cite{cgw05} as an extension to previous well-known action languages for modeling hybrid domains. Then, in \cite{chi13}, {\sl H}
has been used specifically  to model planning and scheduling tasks in hybrid domains, and reasoning is done via CASP language and solver {\sc ezcsp} as in our paper. However, that approach suffers from some shortcomings, which our work overcomes, e.g., it $(i)$ does not take into account PDDL+ as target language, despite it being the standard language for planning in hybrid domains, $(ii)$ does not consider additive fluents, non-linear numerical constraints, and the triggering of processes and events via the must semantics, and $(iii)$ does not include an experimental evaluation of the approach. Another action language {\sl ADP}~\cite{BaralST02} was introduced earlier to allow for the specification of, e.g., actions with duration and continuous effects.  \\ 

Other lines of research that have dealt with extensions involving  primitives for dealing with continuous change, processes, and (macro-)events include the Event Calculus in, e.g., \cite{Evans90,Shanahan90,MillerS96,CervesatoM00}.



\section{Conclusions}\label{sec:concl}

PDDL+ and CASP languages extend PDDL and ASP to reason with mixed discrete-continuous dynamics. PDDL+ is the standard language for the automated planning community, with a number of interesting domains being represented as PDDL+ models.
In this paper, we have presented a new approach for solving PDDL+ problems by means of an encoding into CASP problems, and extension to the {\sc ezcsp} solving architecture for planning in hybrid domains. Our solution can deal with both linear and non-linear variants of the domains. An experimental analysis, performed on well-known PDDL+ domains, involving some variants of our approach, other CASP solvers and PDDL+ planners on two reasoning tasks, showed the viability of our approach.

\bibliography{101-biblio-mb,102-biblio-dm,103-biblio-mm}

\end{document}